\documentclass{article}
\usepackage{amssymb}
\usepackage{amsmath}
\usepackage{graphicx}
\usepackage{placeins}
\usepackage{enumitem}
\usepackage{subcaption}
\usepackage{wrapfig}
\usepackage{booktabs}
\usepackage{multirow}
\usepackage{booktabs}
\usepackage{xcolor}
\usepackage{soul}
\usepackage{tabularx}
\usepackage{makecell}
\usepackage{array}
\usepackage{amssymb}
\definecolor{AnyOrange}{RGB}{236,136,52}
\newcommand{\anytxt}[1]{\textcolor{AnyOrange}{#1}}
\newcommand{\anycheck}{\textcolor{AnyOrange}{$\checkmark$}}
\usepackage[preprint]{corl_2026} % Use this for the initial submission.

\title{Any2Any: Efficient Cross-Embodiment Transfer for Humanoid Whole-Body Tracking}

\author{
Ming Yang,
Tao Yu$^{\dagger*}$,
Feng Li,
Hua Chen\\
LimX Dynamics\\
$^{\dagger}$Project Lead \quad
$^{*}$Corresponding Author
}

\begin{document}
\maketitle
\begin{figure}[h]
    \centering
    \includegraphics[width=0.8\linewidth]{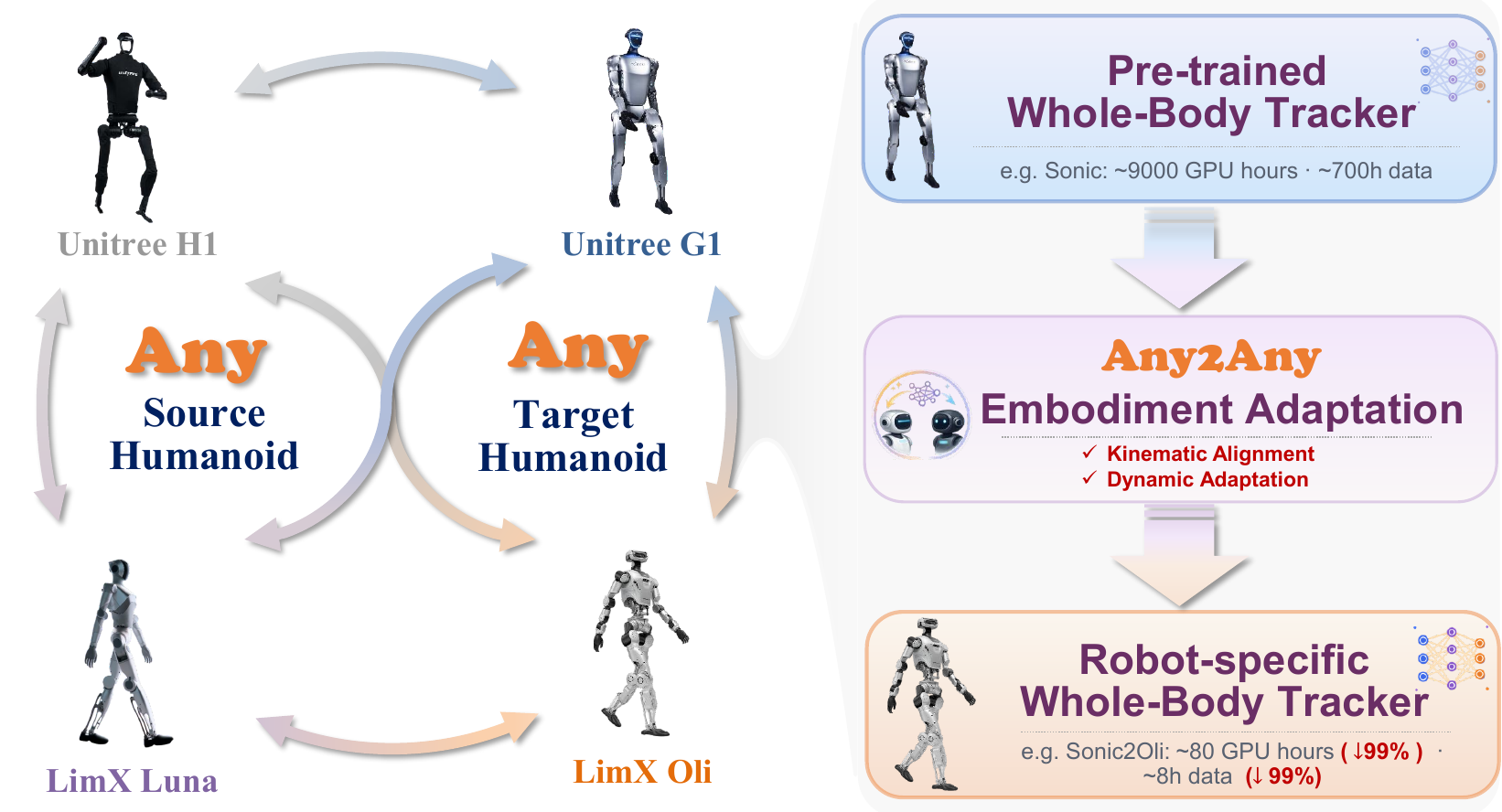}
    \caption{
Illustration of \textsc{Any2Any}. 
A pretrained whole-body tracker (WBT) learned on specific humanoid can be efficiently transferred to another humanoid platform through the proposed \textsc{Any2Any}. 
For example, \textsc{Gear-Sonic}~\cite{luo2025sonic}, a large-scale pretrained WBT, can be adapted to a target robot LimX Oli using only a small fraction of the original training compute and data.
}
    \label{fig:placeholder}
\end{figure}  
% , which bridges the source--target embodiment gap by combining kinematic alignment and dynamics adaptation, ，producing a robot-specific WBT with substantially reduced training cost. 
% Hapfigure
%===============================================================================

\begin{abstract}
Whole-body tracking (WBT) models have become a key foundation for humanoid robots, enabling them to imitate diverse motions with high fidelity. Training such models from scratch requires large-scale data and computation, making rapid deployment on new humanoid platforms costly. This raises a natural question: \textit{Can pretrained WBT models transfer across embodiments with minimal adaptation?}  To answer this question, we propose \textsc{Any2Any}, a paradigm that efficiently transfers an existing robot-specific WBT policy to a new humanoid embodiment with only a small amount of data and compute. Any2Any first performs \mbox{\textit{kinematic alignment}} between source and target humanoids, aligning their input and output spaces so that the pretrained source policy can be meaningfully reused on the target embodiment. Any2Any then performs \textit{dynamics adaptation}, inserting lightweight parameter-efficient fine-tuning (PEFT) components into only the most dynamics-sensitive modules to preserve the source policy's behavioral priors while correcting for the target robot. Extensive experiments on multiple humanoid platforms and pretrained backbones show that Any2Any substantially accelerates convergence and reduces training cost compared with training from scratch, while achieving competitive or superior tracking performance. Notably, using only 1\% of the compute and data required for full training, Any2Any successfully transfers Sonic models pre-trained on Unitree G1 to LimX Oli and LimX Luna. These results suggest that pretrained robot-specific WBT policies can be efficiently reused across embodiments, providing a scalable path toward  deploying humanoid whole-body control on new robots. More results and videos are available on our project page: \url{https://any2any.top/}.
% The \textsc{Sonic}-to-any-humanoid transfer skill is available at:
% \url{https://github.com/YMing2001/Any2Any.git}.
\end{abstract}


%===============================================================================

\section{Introduction}

% Humanoid robots have emerged as a prominent platform for embodied intelligence, owing to their human-like morphology that naturally affords versatility, seamless adaptation to human-centered environments, and intuitive human-robot interaction. 
% A central capability for such robots is whole-body tracking (WBT), where the controller must coordinate legs, torso, arms, and head to reproduce diverse full-body motions while maintaining balance.

% Humanoid robots have emerged as a prominent platform for embodied intelligence, owing to their human-like morphology that naturally affords versatility, seamless adaptation to human-centered environments, and intuitive human-robot interaction. 
% Behavior foundation models (BFMs) are emerging as a new paradigm for humanoid control.
% Instead of training a separate controller for each task, a BFM is pretrained on large-scale behavior data to acquire reusable primitive skills and broad behavioral priors.
% For humanoid robots, whole-body tracking (WBT) provides a particularly scalable interface, where diverse human or robot motions are converted into reference trajectories and the policy learns to reproduce them while maintaining balance.
% In this sense, a large-scale WBT policy can be viewed as a BFM-style whole-body motor prior for humanoid robots, serving as a reusable behavioral backbone for teleoperation, motion imitation, and downstream loco-manipulation.

Behavior foundation models (BFMs)~\cite{pirotta2024fast,yuan2025survey} are emerging as a promising paradigm for humanoid control. By pretraining on large-scale behavior data, BFMs aim to acquire reusable motor skills and broad behavioral priors that can be rapidly adapted to downstream tasks \cite{cetin2024finer, li2025bfm}. For humanoid robots, whole-body tracking (WBT)~\cite{gu2026humanoid, cheng2024expressive} is a natural BFM-style control interface: the policy learns to reproduce diverse full-body reference motions while coordinating legs, torso, arms, and head to maintain balance. A large-scale WBT policy can therefore serve as a reusable whole-body motor prior for teleoperation~\cite{he2024omnih2o, ze2025twist}, motion imitation~\cite{chen2025gmt,luo2025sonic}, and downstream loco-manipulation.

Recent humanoid WBT systems have advanced rapidly, scaling up in data, model size, and motion coverage. TWIST~\cite{ze2025twist} and GMT~\cite{chen2025gmt} demonstrate that end-to-end WBT policies can cover diverse whole-body motions through reinforcement learning, behavior cloning, teacher-student training, adaptive sampling, and mixture-of-experts designs. SONIC~\cite{luo2025sonic} further frames motion tracking as a scalable foundation task, scaling the policy from 1.2M to 42M parameters, using over 100M motion frames, and reporting 9k GPU hours of training. HoloMotion~\cite{chen2026holomotion} trains a large-scale whole-body control foundation model on extensive motion data, while OmniXtreme~\cite{wang2026omnixtreme} scales tracking to high-dynamic, extreme motions without sacrificing fidelity as the motion library grows. These results mark important progress toward general humanoid motor priors, but they also expose a practical barrier: reproducing such natural and robust whole-body behaviors requires large-scale motion data, massively parallel simulation infrastructure, and substantial compute budgets.

Beyond training cost, embodiment dependence~\cite{gupta2022metamorph,yang2025multiloco} remains a fundamental limitation of current large-scale WBT policies.
Such policies are tightly coupled to the source robot's morphology and actuator configuration. Consequently, even between similar humanoids, structural and dynamic differences make direct deployment unreliable, while full-policy fine-tuning tends to overwrite the source behavioral prior. Existing cross-embodiment methods mainly 
address this issue by training universal controllers over many embodiments with morphology randomization or multi-embodiment data~\cite{xue2026xhugwbc,liu2025locoformer,yang2025multiloco}, 
or by scaling robot foundation models with embodiment-aware representations, unified action spaces, post-training, or expert routing~\cite{bjorck2025groot,luo2026being,bai2026hex}. 
However, these approaches typically require large-scale multi-embodiment datasets, broad morphology 
randomization, or expensive large-scale pretraining from scratch, and they mostly address locomotion or manipulation rather than high-fidelity whole-body tracking. In contrast, we study a complementary post-training problem: given a pretrained robot-specific WBT policy, can we efficiently adapt it to another humanoid with limited data and compute?

% \cite{xue2026xhugwbc,lin2025hzero,yang2025multiloco}

A long-standing principle in humanoid control is to decouple kinematics from dynamics: classical pipelines typically plan a kinematic reference and track it with a hierarchical whole-body controller \cite{kim2018prioritized, chignoli2021mit}, while learning-based whole-body tracking (WBT) policies often encode retargeted reference motions~\cite{araujo2025retargeting} separately from the dynamics-aware control stream~\cite{luo2025sonic}. This decoupling reveals the structure of cross-embodiment transfer. A source policy trained on one humanoid differs from a target humanoid along two distinct axes. \textit{Kinematically}, the two robots may have different joint counts, link geometries, and observation/action layouts, making the pretrained policy structurally incompatible with the target embodiment. \textit{Dynamically}, they differ in mass distributions, inertias, actuator responses, and contact behaviors, so even after interface alignment, the source policy may still produce suboptimal or incorrect actions on the target. Bridging this dynamics gap requires parameter updates, but directly fine-tuning the full policy risks overwriting the very behavioral prior that makes it valuable.
This kinematics-dynamics separation is also reflected in modern WBT network designs, which commonly adopt a two-stream structure consisting of a reference motion encoder and an action decoder~\cite{ze2025twist,chen2025gmt,luo2025sonic,zeng2025behavior,zhu2026clot,chen2026holomotion}. The encoder primarily extracts motion-related, largely kinematic features that are more transferable across embodiments, whereas the decoder is more tightly coupled to embodiment-specific dynamics. We therefore hypothesize that embodiment changes affect different components of the WBT prior unevenly, i.e. global motor skills such as balance and inter-limb coordination may remain broadly reusable, while embodiment-sensitive modules require adaptation. This motivates a localized adaptation strategy, where lightweight PEFT components~\cite{ding2023peft,li2021prefix} such as LoRA~\cite{hu2022lora} and adapters~\cite{houlsby2019adapter} are inserted only into modules with the largest source-target discrepancy, while the remaining parameters are frozen to preserve the source prior. This leads to three key questions: \textit{Can localized fine-tuning adapt to the target embodiment without destroying the source prior?} \textit{How should the kinematic mismatch be resolved?} \textit{Which modules should be adapted to maximize cross-embodiment transfer?}

% aspects

% To answer this question, we propose Any2Any, a parameter-efficient post-training framework for cross-humanoid WBT. We instantiate the setting by transferring a high-performance Oli WBT policy to Oli Lite. Any2Any inserts lightweight trainable modules into selected components of the pretrained policy while freezing most source parameters. We systematically evaluate multiple PEFT strategies, including LoRA variants, prefix tuning, and adapters, and compare them with full fine-tuning and training-from-scratch baselines. Our contributions are threefold:
To answer these questions, we propose \textsc{Any2Any}, a parameter-efficient post-training framework that decomposes cross-embodiment WBT transfer into two complementary steps, \textit{kinematic alignment} and \textit{dynamics adaptation}. 
Our contributions are threefold:
\begin{enumerate}[leftmargin=12pt]
% \item To the best of our knowledge, we present the first systematic study demonstrating that humanoid WBT policies pretrained on a single specific robot can be effectively transferred to other humanoid embodiments through principled handling of the kinematic and dynamics mismatches, establishing cross-embodiment WBT transfer as a viable problem setting.
\item We demonstrate that humanoid WBT policies pretrained on a single source robot can be effectively transferred to other humanoid embodiments through principled handling of the kinematic and dynamics mismatches. To the best of our knowledge, this constitutes the first systematic study of cross-embodiment transfer for humanoid WBT.
\item We propose \textsc{Any2Any}, a cross-embodiment post-training framework for humanoid WBT that decomposes transfer into two stages: \textit{kinematic alignment}, which resolves the structural kinematic mismatch between source and target robots, and \textit{dynamics adaptation}, which adapts only the dynamics-sensitive modules while freezing the rest of the pretrained backbone.
% \item We systematically study \textit{where} and \textit{how} to insert lightweight PEFT components for cross-embodiment WBT adaptation, comparing different insertion positions and PEFT mechanisms to provide empirical guidance.
\item We validate {Any2Any} on five source-target transfer pairs spanning two pretrained WBT backbones and four target humanoid embodiments, achieving successful transfer with only $\sim$1\% of the compute and data of  training from scratch, and deploy the adapted policies on real hardware across multiple downstream tasks.
\end{enumerate}

\section{Related Works}

% \subsection{Humanoid Whole-Body Tracking}

% V1
% Learning-based humanoid whole-body control has progressed from task-specific skill learning toward large-scale motion tracking and reusable motor priors \cite{peng2018deepmimic,peng2021amp,ze2025twist,chen2025gmt,luo2025sonic}. Early works such as DeepMimic and AMP showed that reinforcement learning with motion references or adversarial motion priors can produce natural full-body behaviors \cite{peng2018deepmimic,peng2021amp}. Recent real-humanoid systems extend this paradigm to whole-body tracking and teleoperation. TWIST formulates teleoperation as real-time whole-body retargeting and tracking, while CLONE improves long-horizon teleoperation by addressing decoupled upper/lower-body control and open-loop drift \cite{ze2025twist,li2025clone}. HuB studies balance-intensive motions and highlights reference errors, morphology mismatch, and sim-to-real gaps as key obstacles \cite{zhang2025hub}. SONIC further scales WBT with larger policies and motion datasets, treating motion tracking as a foundation task for humanoid control \cite{luo2025sonic}.

% These methods demonstrate the promise of WBT as a general humanoid motor interface. However, they primarily focus on training strong policies for a given robot or scaling WBT from scratch. In contrast, Any2Any studies how to reuse a trained WBT policy and adapt it efficiently to a new humanoid embodiment.

% V2
\paragraph{Humanoid Whole-Body Tracking.}
Humanoid control has progressed from task-specific motion imitation to large-scale whole-body tracking (WBT). 
DeepMimic~\cite{peng2018deepmimic} demonstrates that physics-based reinforcement learning can imitate reference motions and produce natural full-body behaviors, while PHC~\cite{luo2023phc} improves scalable humanoid motion tracking from large-scale unstructured human motion data. 
Recent studies further extend WBT to real humanoids, including teleoperation~\cite{ze2025twist}, general motion tracking~\cite{chen2025gmt}, long-horizon closed-loop control~\cite{li2025clone}, and balance-intensive motions~\cite{pan2025agility}.
SONIC scales this paradigm with larger policies and motion datasets, showing the potential of WBT as a foundation-level motor interface~\cite{luo2025sonic}. 
However, these policies are still trained and tuned for a specific robot embodiment~\cite{pan2025agility,sun2026mosaic,wang2026omnixtreme}. 
Their observation layout, action space, reward design, and dynamics randomization are tightly coupled to one morphology. 
In contrast, \textsc{Any2Any} studies how to transfer an existing WBT policy and efficiently adapt it to a different humanoid embodiment.

\paragraph{Cross-Embodiment Robot Learning.}
Cross-embodiment learning aims to share control knowledge across robots with different morphologies. Existing methods commonly train generalist controllers over multiple embodiments using morphology randomization~\cite{xue2026xhugwbc}, topology-aware architectures~\cite{lin2025hzero}, unified observation-action spaces~\cite{liu2025locoformer}, or residual adaptation modules~\cite{yang2025multiloco}. In robot foundation models, cross-embodiment generalization is often addressed through
large-scale pretraining~\cite{o2023openx,octo2024},
embodiment-aware architectures~\cite{bjorck2025groot},
and unified action/state representations with expert routing~\cite{luo2026being,bai2026hex}. These methods are powerful, but they typically require multi-embodiment datasets, broad robot coverage, and expensive training from scratch. This is difficult for WBT, where high-quality specialist policies already require substantial data and compute. \textsc{Any2Any} addresses a complementary setting: given one pretrained WBT specialist, we transfer it to a target humanoid through kinematic alignment and lightweight policy adaptation, without rebuilding a multi-robot generalist model.

\paragraph{Parameter-Efficient Fine-tuning for Robotics.}
A direct way to adapt a pretrained model is full fine-tuning, which updates all model parameters for the target domain~\cite{devlin2019bert}. Parameter-efficient fine-tuning instead freezes most pretrained weights and 
updates only a small set of task-specific parameters. Adapter tuning inserts trainable bottleneck modules~\cite{houlsby2019adapter}; Prefix-Tuning optimizes continuous prefix tokens while keeping the backbone frozen~\cite{li2021prefix}; LoRA represents weight updates with low-rank factors~\cite{hu2022lora}.
Recent works have begun to adapt PEFT techniques to robotic foundation models, enabling efficient specialization to downstream tasks and 
embodiments without retraining the entire  policy~\cite{kim2025finetuning,wang2025vlaadapter}. 
However, adapting a closed-loop humanoid WBT policy is fundamentally different from adapting a static prediction model: small parameter changes can propagate through contact dynamics, balance regulation, and action feedback, leading to non-trivial shifts in the resulting state distribution. 
It therefore remains unclear which PEFT mechanism and where PEFT components should be inserted, and to what extent the motion priors encoded in an existing WBT policy can be reused on a structurally different humanoid.
In this work, \textsc{Any2Any} aims to fill this gap by introducing a systematic paradigm tailored for humanoid whole-body tracking, which transfers pretrained WBT policy to new humanoids with only a small amount of adaptation data and compute.

\section{Method}

\begin{figure}
    \centering

    \includegraphics[width=\linewidth]{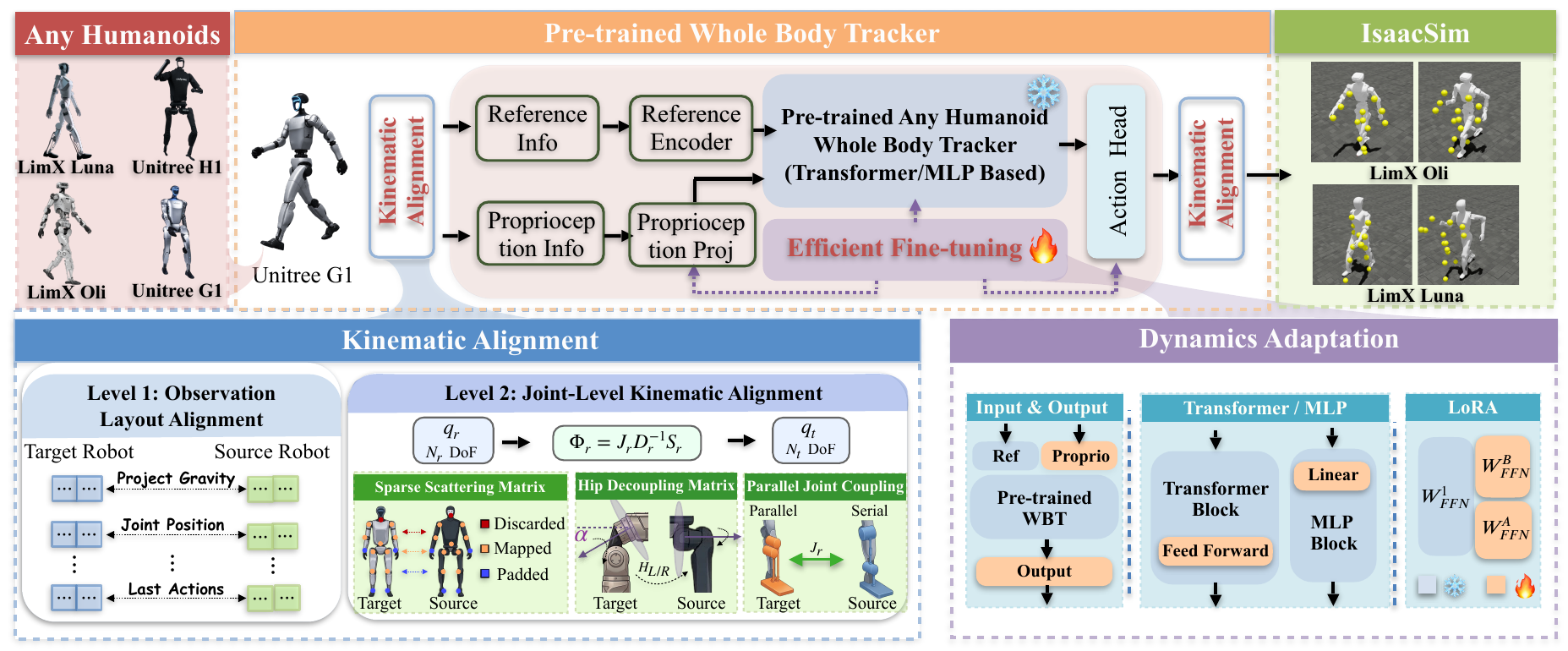}
    \caption{\textbf{Architecture of Any2Any.} The proposed framework adapts a pretrained whole-body tracker to arbitrary humanoid embodiments by combining \textbf{Kinematic Alignment}, which maps observations and actions across different robot morphologies, with \textbf{Dynamics Adaptation}, which efficiently fine-tunes lightweight modules to account for target-specific dynamics.
 }
    \label{fig:pipeline}
\end{figure}

\subsection{Problem Formulation}
We formulate humanoid whole-body tracking (WBT) as a Markov Decision Process (MDP) $\mathcal{M} = (\mathcal{X}, \mathcal{A}, \mathcal{P}, r, \gamma)$, with state space $\mathcal{X}$, action space $\mathcal{A}$, transition kernel $\mathcal{P}$, reward $r$, and discount $\gamma$. For a given robot embodiment, $\mathcal{X}$ and $\mathcal{A}$ are determined by its morphological properties (number of actuated joints, kinematic structure, link masses, inertias, actuator gains) and by the specific tracking task.

Let $\mathcal{E}_{\mathcal{S}}$ denote a source robot equipped with a pretrained WBT policy $\pi_{\theta_{\mathcal{S}}}: \mathcal{X}_{\mathcal{S}} \to \mathcal{A}_{\mathcal{S}}$, where $\theta_{\mathcal{S}}$ is its full parameter set. For a target robot $\mathcal{E}_{\mathcal{T}}$, $\mathcal{X}_{\mathcal{S}} \neq \mathcal{X}_{\mathcal{T}}$ and $\mathcal{A}_{\mathcal{S}} \neq \mathcal{A}_{\mathcal{T}}$ due to differences in joint configuration and degrees of freedom. Re-training the full parameter set $\theta_{\mathcal{S}}$ on $\mathcal{E}_{\mathcal{T}}$ is computationally expensive and risks catastrophic forgetting of the transferable motor priors encoded in $\theta_{\mathcal{S}}$.

We instead freeze $\theta_{\mathcal{S}}$ and learn a small target-specific adaptation $\Delta\theta_{\mathcal{T}}$ with $|\Delta\theta_{\mathcal{T}}| \ll |\theta_{\mathcal{S}}|$. The resulting target policy is
\begin{equation}
\pi_{\theta_{\mathcal{T}}}(a_t \mid s_t) \;=\; \pi_{\theta_{\mathcal{S}} \,\oplus\, \Delta\theta_{\mathcal{T}}}(a_t \mid s_t),
\end{equation}
where $\oplus$ denotes the parameter-efficient injection of $\Delta\theta_{\mathcal{T}}$ into selected modules of the frozen backbone. The adaptation is trained under the standard expected-return objective
\begin{equation}
\max_{\Delta\theta_{\mathcal{T}}} \;\; \mathbb{E}_{\pi_{\theta_{\mathcal{T}}}} \!\left[ \sum_{t=0}^{T} \gamma^t \, r(s_t, a_t) \right].
\end{equation}

Different PEFT mechanisms such as LoRA, Adapter, and Prefix Tuning realize the operator $\oplus$ at different locations within the network, yielding distinct trade-offs in parameter efficiency, training stability, and tracking quality. Identifying which positions and mechanisms yield the most effective cross-embodiment WBT adaptation under limited data and compute is the focus of this work.

\subsection{Pretrained Whole Body Tracking Policy Structure}

We adopt an actor--critic framework trained with Proximal Policy Optimization (PPO) for motion imitation. 
The actor $\pi_\theta$ outputs target joint position offsets, while the critic $V_\phi$ estimates state values using privileged observations (e.g., base velocity, contact forces, body mass distribution) that are available only during training. 
To demonstrate the generalizability of the proposed method, we instantiate the actor with three representative policy backbones that share a unified observation formulation but differ in how they encode and aggregate the inputs. The first two backbones are trained under the Oli whole-body tracking setting and are collectively referred to as \textbf{Oli-WBT}.

\textbf{Unified Observation Formulation.}
At each control step $t$, the policy input is organized into two components: proprioceptive feedback and reference motion information.
\begin{itemize}[leftmargin=8pt]
    \item \textbf{Proprioception} ($\phi_t \in \mathbb{R}^{d_p}$):
    the robot-centric state, including base angular velocity, projected gravity, joint positions, joint velocities, and the previous action.
    These observations provide feedback about the current physical state and recent control history of the robot.
    \item \textbf{Reference} ($g_t \in \mathbb{R}^{d_r}$):
    the target motion state extracted from the reference trajectory, which provides task-level guidance for motion imitation.
\end{itemize}
% To incorporate temporal context, all backbones take as input a sliding window of $H{+}1$ timesteps, i.e., the current step and the $H$ preceding ones.
% To incorporate temporal context, all backbones maintain a sliding window of $H$ historical timesteps. 
% Thus, at each decision step, the policy receives the observation history
% \begin{equation}
%     \{(\phi_{t-H}, g_{t-H}), \ldots, (\phi_t, g_t)\}.
% \end{equation}

\textbf{MLP Backbone.}
The MLP policy flattens multi-modal observations over $H+1$ timesteps into a vector of dimension $(H+1)(d_p+d_r)$, which is then processed by an $L$-layer MLP with GELU activations to predict actions. 
The critic uses the same architecture with privileged observations as input. 
This simple yet effective design is widely adopted in recent humanoid whole-body tracking works~\cite{he2024omnih2o, cheng2024expressive}.

\textbf{Transformer Backbone.}
The Transformer policy models temporal dependencies by projecting each modality into a shared $d$-dimensional embedding space and processing the resulting sequence with $N$ causal Transformer Decoder blocks. 
The action is decoded from the current-timestep representation, while the critic follows the same architecture with privileged observations.

\textbf{Sonic Backbone.}
We further adopt Sonic~\cite{luo2025sonic} as a representative large-scale 
humanoid motion tracking architecture. In our experiments, we employ its Robot 
Motion Encoder, FSQ bottleneck, and dynamics decoder modules for training and evaluation. 
More architectural details can be found in~\cite{luo2025sonic}.

\subsection{Cross-Embodiment Adaptation}

% Since the source and target robots may differ in degrees of freedom $(n_j^S \neq n_j^T)$, their observation and action spaces cannot be directly shared. Before applying any PEFT method, a kinematic Calibration is introduced to align the joint-level representation between the two embodiments. This mapping is shared across all adaptation strategies.

Since the source and target robots may differ in their degrees of freedom $(n_j^S \neq n_j^T)$, their observation and action spaces cannot be directly shared. Before applying any PEFT method, we introduce a kinematic alignment step to construct a unified joint-level representation across embodiments.

\subsubsection{Kinematic Alignment}
To bridge the embodiment gap between the source and target robots before any parameter adaptation, we introduce a two-level kinematic alignment module that operates on the policy's input and output streams (Fig.~\ref{fig:pipeline}). 
The first level aligns the \emph{global observation layout}, while the second level aligns the \emph{joint-level kinematic representation}. 
Together, they guarantee that the frozen source policy always receives observations with a consistent semantic layout and a compatible joint ordering, while the target robot still executes actions in its own control convention.

\paragraph{Level 1: Observation Layout Alignment.}
Different humanoid platforms often organize their observation vectors in different orders, even when the underlying entries are semantically equivalent.
At this level, we rearrange the target robot's observation components, such as base states, reference motion features, proprioceptive states (e.g., projected gravity, joint positions), and action history, to match the input convention of the pretrained source policy, so that each term lands at the position expected by the frozen backbone.

\paragraph{Level 2: Joint-Level Kinematic Alignment.}
The second level aligns the internal ordering and geometry of joint-related variables. 
Consider a family of robots $\mathcal{R}=\{r_1, r_2, \dots\}$ that share the same humanoid topology but differ in joint count, where each robot $r$ has $N_r$ degrees of freedom. 
We treat the joint layout of the pretrained source robot $r_0$, with $N_{r_0}=T$, as a unified kinematic semantic space, and for every $r \in \mathcal{R}$ we construct a pair of mappings $\Phi_r:\mathbb{R}^{N_r}\!\to\!\mathbb{R}^{T}$ and $\Phi_r^{+}:\mathbb{R}^{T}\!\to\!\mathbb{R}^{N_r}$ between the target and source joint spaces. On the joints shared by both robots, $\Phi_r^{+}$ inverts $\Phi_r$, so that $\Phi_r^{+}\Phi_r$ is the identity on these matched target joints and reduces to $I_{N_r}$ when every target joint has a source counterpart. 
The mapping $\Phi_r$ is composed of three structured components that progressively account for joint-level discrepancies between the two embodiments.

\textbf{(i) Sparse Scattering Matrix.}
Joint correspondence between the target and source robots is specified by a partial injection $\pi_r:\{0,\dots,N_r-1\}\rightharpoonup\{0,\dots,T-1\}$ defined on the target joints that have a source counterpart, which induces a sparse scattering matrix $S_r\in\{0,1\}^{T\times N_r}$ with $(S_r)_{ij}=\mathbf{1}[\pi_r(j)=i]$. 
Each matched target joint yields a column with a single nonzero entry, so $S_r$ scatters target values into the source layout: source positions without a target counterpart become zero rows (zero-padding), while surplus target joints without a source counterpart lie outside the domain of $\pi_r$ and become zero columns (dropped).

\textbf{(ii) Hip Decoupling Matrix.}
The source embodiment contains an inclined hip-pitch axis, which induces a coupling 
between the corresponding hip coordinates. To compensate for this structural difference, 
we introduce a hip decoupling matrix $D_r \in \mathbb{R}^{T\times T}$. 
Specifically, $D_r$ is initialized as a $T\times T$ identity matrix, and its left and 
right hip submatrices are replaced by the following decoupling blocks:
\begin{equation}
H_{\mathrm{L/R}} \;=\;
\begin{pmatrix}
\cos\alpha & 0 \\
\mp\sin\alpha & 1
\end{pmatrix},
\end{equation}
where the opposite signs account for the mirrored hip-axis orientations on the left and 
right legs. In this way, $D_r$ acts only on the hip-related coordinates and leaves all 
other joints unchanged.

Combining the scattering matrix $S_r$ and the hip decoupling matrix $D_r$ gives the 
aligned forward and inverse mappings
\begin{equation}
\Phi_r \;=\; D_r^{-1}\, S_r, 
\qquad
\Phi_r^{+} \;=\; S_r^{\top}\, D_r .
\end{equation}
Here, $S_r$ first scatters target-robot joint quantities into the source joint layout, 
and $D_r^{-1}$ further converts the hip coordinates into the source-aligned convention. 
Conversely, $D_r$ maps the policy output back from the source-aligned convention, and 
$S_r^\top$ gathers the corresponding entries for the target robot. 
% Since each column of $S_r$ has exactly one nonzero entry, the mappings satisfy
% \begin{equation}
% \Phi_r^{+}\Phi_r
% =
% S_r^{\top} D_r D_r^{-1} S_r
% =
% S_r^{\top} S_r
% =
% I_{N_r}.
% \end{equation}

Structured joint-related terms in the observation, including reference motion, joint 
position observations, and action history, are mapped to the source space via 
$\widetilde{\mathbf{q}}_r=\Phi_r\,\mathbf{q}_r$. 
The same rule is applied to the action space: the policy output $\widetilde{\mathbf{a}}$ 
produced under the source-aligned joint convention is converted back to the target 
robot's actuated joints through $\mathbf{a}_r=\Phi_r^{+}\,\widetilde{\mathbf{a}}$ for 
execution.

\textbf{(iii) Parallel Joint Coupling.}
Beyond serial-chain joint correspondence, some target robots contain closed-chain mechanisms, such as parallelogram-driven ankles or closed-loop waists, whose actuated joint values do not coincide with their kinematic counterparts in the source serial chain. 
For these joints, we incorporate a closed-chain Jacobian correction $J_r \in \mathbb{R}^{T\times T}$ into the mapping, so that the effective joint-level alignment becomes
\begin{equation}
\Phi_r \;=\; J_r\, D_r^{-1}\, S_r, 
\qquad
\Phi_r^{+} \;=\; S_r^{\top}\, D_r\, J_r^{-1},
\end{equation}
which preserves this relation on the matched joints. 
This term explicitly handles parallel-to-serial discrepancies that purely permutation-based alignment cannot capture.

\medskip
\noindent
Through these two levels of alignment, the frozen source policy operates on a stable kinematic semantic space, while embodiment-specific details, such as joint ordering, hip-axis inclination, and closed-chain coupling, are absorbed by the alignment module itself. 
This consistently reduces the embodiment gap before parameter adaptation, and provides a more reliable basis on which subsequent PEFT methods can specialize the policy to the target humanoid.

\subsubsection{Dynamics Adaptation}

Once kinematic alignment is in place, the remaining gap between the source embodiment 
$\mathcal{S}$ and the target embodiment $\mathcal{T}$ is no longer representational but 
\emph{dynamical}. Here, $\mathcal{S}$ denotes the humanoid embodiment on which the 
whole-body tracker is originally pretrained, while $\mathcal{T}$ denotes the new target 
humanoid embodiment to which the tracker is transferred. After kinematic alignment, we use $q$ to denote the joint state expressed in the shared source-aligned joint convention. The rigid-body dynamics of a 
humanoid admits the standard manipulator-equation form
\begin{equation}
M(q)\,\ddot{q} + C(q,\dot{q})\,\dot{q} + G(q) \;=\; \tau + \tau_{\text{ext}},
\end{equation}
which makes explicit that an embodiment's physical identity is carried by three 
state-dependent terms: the mass matrix $M(q)$, the Coriolis/centrifugal coupling 
$C(q,\dot{q})$, and the gravity loading $G(q)$. Under the same motion target 
$(q, \dot{q}, \ddot{q})$, two humanoids with different dynamics 
$(M_{\mathcal{S}}, C_{\mathcal{S}}, G_{\mathcal{S}})$ and 
$(M_{\mathcal{T}}, C_{\mathcal{T}}, G_{\mathcal{T}})$ require different generalized 
forces to realize that motion, yielding the rigid-body residual
\begin{equation}
\Delta\tau(q,\dot{q},\ddot{q}) \;=\; 
\Delta M(q)\,\ddot{q} \,+\, 
\Delta C(q,\dot{q})\,\dot{q} \,+\, 
\Delta G(q),
\end{equation}
where $\Delta M = M_{\mathcal{T}} - M_{\mathcal{S}}$, 
$\Delta C = C_{\mathcal{T}} - C_{\mathcal{S}}$, and 
$\Delta G = G_{\mathcal{T}} - G_{\mathcal{S}}$. Beyond these rigid-body terms, the 
realized force-motion relation also depends on unmodeled effects such as actuator 
characteristics, joint friction, and contact behavior, which contribute additional 
residual components on top of $\Delta\tau$. Collecting all embodiment-specific physical 
quantities into a single descriptor $\eta_e$, the full cross-embodiment dynamics gap can 
be characterized by
\begin{equation}
\Delta\eta \;=\; \eta_{\mathcal{T}} - \eta_{\mathcal{S}}.
\end{equation}

For humanoids of similar topology and scale, $\Delta\eta$ is structurally low-dimensional: only a limited subset of physical quantities differs between the two robots, and the rigid-body part $(\Delta M, \Delta C, \Delta G)$ is fully parameterized by per-link inertial differences, whose number scales with the link count rather than with the size of the full backbone $\theta_{\mathcal{S}}$. Once kinematic alignment has removed the dominant structural mismatch, $\Delta\eta$ becomes the primary remaining source of embodiment-dependent policy error. The pretrained policy has already captured the task structure (reference tracking, balance, inter-limb coupling) in a source-aligned representation; what remains is a compact, embodiment-specific correction whose capacity, we hypothesize, should scale with $\Delta\eta$ rather than with $\theta_{\mathcal{S}}$. We therefore adapt the target robot through a low-rank correction rather than full-parameter fine-tuning, understanding it as a learned representation of the source--target dynamics residual rather than a generic perturbation of the source weights.

This motivates a strongly asymmetric adaptation: we freeze the pretrained parameters $\theta$ and learn an embodiment-specific low-rank correction. Concretely, for each adapted linear projection $W \in \mathbb{R}^{d_{\text{out}} \times d_{\text{in}}}$ in the policy, we apply a LoRA decomposition
\begin{equation}
W' \;=\; W \;+\; B A, \qquad A \in \mathbb{R}^{k \times d_{\text{in}}},\; B \in \mathbb{R}^{d_{\text{out}} \times k},
\end{equation}
with rank $k \ll \min(d_{\text{in}}, d_{\text{out}})$. Only $\{A, B\}$ are trained per target embodiment, while $W$ remains shared with the source. The rank $k$ controls how much capacity is allocated to absorbing the cross-embodiment dynamics gap, providing a tunable trade-off between transfer fidelity and adaptation cost.

Together with the kinematic alignment, this design factorizes the cross-embodiment gap into a fixed structural component absorbed by the alignment, and a learned dynamical component absorbed by LoRA, keeping the source policy's behavioral prior intact while allowing each new embodiment to be specified by a small set of low-rank factors. Concretely, we inject LoRA into the proprioceptive input projection and the actor and critic linear layers, while keeping the reference-motion encoder frozen.
\subsection{Training Procedure}
All policies are trained with PPO in Isaac Lab. To isolate the effect of cross-embodiment adaptation, we keep the action space, observations, reward formulation, PPO hyperparameters, reference-motion sampling, and domain randomization protocol identical to those of the corresponding source pretraining setup, only the components introduced by our kinematic alignment and dynamics adaptation differ.

\section{Experimental Results}
% 为了解答以下问题，我们进行了实验：
% 1. 什么样的PEFT方法对于Cross-Embodiment是有效的？
% 2. CrossTuning在跨机器人上能做到什么样的泛化效果？
% 3. 能否基于成熟的开源模型，如Sonic，在更小的算力，更短的时间上，收敛到一个较好的效果？
% In this section, our experimental evaluations are designed to thoroughly assess the proposed \textsc{Any2Any} framework in cross-embodiment whole-body control scenarios. Specifically, we structure our experiments and ablation studies to address the following three principal research questions:

Our experiments are designed to test the central hypothesis of Any2Any: a robot-specific pretrained whole-body tracking policy can be efficiently reused on a new humanoid if the cross-embodiment gap is decomposed into a structural kinematic mismatch and a compact dynamics residual.

\begin{itemize}[leftmargin=1.5em, itemsep=4pt]
    \item \textbf{Q1:} Can the \textsc{Any2Any} successfully transfer diverse pretrained whole-body tracking policies to novel robotic platforms with varying morphologies?

    % \item \textbf{Q2:} Compared to training policies from scratch, what are the comparative advantages of \textsc{Any2Any} concerning its sensitivity to training data volume (sample efficiency) and overall computational efficiency?

\item \textbf{Q2:}
How much target-side data and compute does Any2Any save compared with training a target specialist from scratch?

    \item \textbf{Q3:} What are the specific contributions and roles of the individual algorithmic components within the \textsc{Any2Any} paradigm ?

\end{itemize}
% 1. Cross Embodiment：Lite   不同的Finetuning 方法之间的对比
% 2. Cross Embodiment：Lite   Ablation：LoRA的不同模式之间的对比

% 3. Application：Cross-Embodiment、dataset不同量级下的泛化比较、dataset内的追踪效果。  Lite真机？
% 4. Sonic的finetune实验：真机

\subsection{Experimental Protocol}
\subsubsection{Cross-Embodiment Transfer Benchmark}

% We build our cross-embodiment benchmark upon the LimX Oli, a full-size humanoid robot with 31 degrees of freedom (DoF). Based on this platform, we first pretrain a foundational Whole-Body Tracking (WBT) policy using both Transformer and MLP backbones. The pretraining stage leverages over 500 hours of motion data and requires substantial computational resources and prolonged training time to acquire robust whole-body motion priors.

% To evaluate the generality of our cross-embodiment adaptation framework, we transfer pretrained WBT policies across multiple humanoid platforms with distinct morphologies and body scales, including the Unitree G1, Unitree H1, LimX Oli, and LimX Luna. Specifically, we consider two types of source policies: (1) our self-pretrained Transformer-based WBT policy on LimX Oli Humanoid (\textbf{Oli-Pretrained WBT}), and (2) NVIDIA open-sourced Gear-Sonic, a large-scale foundational WBT policy originally trained for the Unitree G1 (\textbf{Sonic}). Starting from these pretrained policies, we fine-tune them on target embodiments using only a small amount of adaptation compute (4 NVIDIA H100/A100 GPUs), demonstrating that our framework is capable of adapting arbitrary pretrained WBT policies to arbitrary humanoid embodiments.

We build our cross-embodiment benchmark on four humanoid platforms spanning two
morphological families and $19$--$31$ degrees of freedom (DoF): LimX
Oli~\cite{limx_oli} (31-DoF), LimX Luna~\cite{limx_luna} (27-DoF), Unitree
G1~\cite{unitree_g1} (29-DoF), and Unitree H1~\cite{unitree_h1} (19-DoF),
summarized in Table~\ref{tab:robots}. Beyond differences in DoF and body scale,
the two families also differ structurally: the LimX robots use an inclined
hip-pitch axis (Oli $25^{\circ}$, Luna $30^{\circ}$) and closed-chain
parallelogram ankle and waist actuation, whereas the Unitree robots use
orthogonal hip axes and fully serial chains. These structural gaps are exactly
what the kinematic alignment of \textsc{Any2Any} is designed to absorb ($S_r$
for the DoF mismatch, $D_r$ for the inclined hip-pitch, and $J_r$ for the
closed-chain actuation), making the benchmark a direct stress test of
cross-embodiment transfer.

\begin{table}[t]
\centering
\scriptsize
\setlength{\tabcolsep}{4pt}
\renewcommand{\arraystretch}{1.25}
\caption{\textbf{Cross-embodiment benchmark robots.} DoF is reported as
leg/arm/waist/head; structural gaps and their role in the kinematic alignment
are discussed in the text.}
\label{tab:robots}
\begin{tabular*}{\linewidth}{@{\extracolsep{\fill}}lccccccl@{}}
\toprule
\textbf{Robot} & \makecell{DoF\\{\scriptsize(l/a/w/h)}} & \makecell{Joint\\conv.}
& \makecell{Incl.\,hip\\$\to D_r$} & \makecell{Closed-ch.\\$\to J_r$}
& Height & Mass & Transfer role \\
\midrule
LimX Oli   & 31\,(12/14/3/2)             & Serial$^{\dagger}$ & $\checkmark$\,$25^{\circ}$ & --$^{\dagger}$ & 1.65\,m & 55\,kg & Src\,$\to$\,G1, H1, Luna;\ \ Tgt\,$\gets$\,Sonic \\
LimX Luna  & 27\,(12/10/3/2)             & Parallel           & $\checkmark$\,$30^{\circ}$ & $\checkmark$   & 1.60\,m & 54\,kg & Tgt\,$\gets$\,Oli-WBT, Sonic \\
Unitree G1 & 29$^{\ddagger}$\,(12/14/3/0) & Serial             & --                         & --             & 1.32\,m & 35\,kg & Src\,$\to$\,Oli, Luna;\ \ Tgt\,$\gets$\,Oli-WBT \\
Unitree H1 & 19\,(10/8/1/0)             & Serial             & --                         & --             & 1.80\,m & 47\,kg & Tgt\,$\gets$\,Oli-WBT \\
\bottomrule
\end{tabular*}

\vspace{2pt}
{\scriptsize $^{\dagger}$Oli is physically parallel (achilles ankle and parallel
waist), but the WBT policy operates on its serialized model with the parallel
mapping resolved at the low level, so $J_r$ is not exercised for Oli.
$^{\ddagger}$$29$-DoF body of the \texttt{g1\_29dof} configuration on which Sonic
is pretrained; hand DoF are excluded from WBT. Heights and masses are nominal
manufacturer values.}
\end{table}

We use two pretrained WBT policies as source backbones.
The first is our self-pretrained \textbf{Oli-WBT}, trained on
LimX Oli with over 500 hours of motion data using both Transformer and MLP
backbones. The second is \textbf{Sonic}~\cite{luo2025sonic}, the open-source
Gear-Soni, a large-scale WBT policy pretrained on Unitree G1, of which we
reuse the Robot Motion Encoder, FSQ bottleneck, and dynamics decoder. From these
backbones we evaluate five transfer instances across two directions: adapting
Oli-WBT to Unitree G1, Unitree H1, and LimX Luna yields \textsc{OliWBT2G1},
\textsc{OliWBT2H1}, and \textsc{OliWBT2Luna}, while adapting Sonic to LimX Oli
, LimX Luna, Unitree H1 yields \textsc{Sonic2Oli},  \textsc{Sonic2Luna} and \textsc{Sonic2H1}. Each adaptation
uses limited target-side compute: 4 NVIDIA A100 GPUs, $\approx$$22.5$\,h of
wall-clock on average, which shows that pretrained WBT priors can be reused across
embodiments at low cost.
% For clarity, we denote each transfer instance by its source and target
% embodiment. Specifically, \textsc{OliWBT2G1}, \textsc{OliWBT2H1}, and
% \textsc{OliWBT2Luna} denote transfers from the Oli-pretrained WBT policy,
% while \textsc{Sonic2Oli} and \textsc{Sonic2Luna} denote transfers from the
% Sonic source policy.

\textbf{Adaptation and evaluation data.}
All adaptation is driven by the AMASS~\cite{mahmood2019amass} motion dataset,
retargeted to each target robot through General Motion
Retargeting~\cite{araujo2025retargeting}; sharing a single adaptation corpus
isolates the effect of the adaptation method across source policies and target
embodiments. For evaluation, we hold out two motion benchmarks chosen to cover
complementary regimes: \textbf{OMOMO}~\cite{li2023omomo}, a whole-body
object-manipulation dataset ($500$ selected clips), and
\textbf{LAFAN}~\cite{harvey2020lafan}, a high-dynamic locomotion dataset (after
removing near-ground, crawling, and retargeting-corrupted clips). As neither is
seen during adaptation, all deployment metrics are reported on these two
held-out sets and thus jointly measure out-of-distribution tracking of
manipulation (OMOMO) and dynamic locomotion (LAFAN).

\subsubsection{Baselines Design}

To evaluate the proposed cross-embodiment adaptation framework, we compare against several representative training and adaptation strategies built on the same PPO pipeline. All methods share identical network architectures, training environments, reward functions, and motion-retargeting pipelines, and crucially are run under the \emph{same compute budget and the same number of training iterations}, so that observed differences reflect the adaptation mechanism rather than the amount of training.

\textbf{Scratch.} Our primary baseline trains the target policy from scratch: it is randomly initialized and optimized with PPO on the retargeted AMASS data, without any pretrained prior. This reflects the conventional per-robot humanoid RL pipeline and is the reference point for adaptation efficiency.

\textbf{Full fine-tuning.} To isolate the value of parameter-efficient adaptation, we also fine-tune the \emph{entire} pretrained backbone in two variants: without kinematic alignment (loading the source weights directly) and with our kinematic alignment; both update all parameters.

\textbf{PEFT mechanisms.} Finally, under the same kinematic alignment, we compare the low-rank adaptation adopted by \textsc{Any2Any} (\textbf{LoRA}~\cite{hu2022lora}) against two alternative parameter-efficient mechanisms, \textbf{Adapter}~\cite{houlsby2019adapter} and \textbf{Prefix-Tuning}~\cite{li2021prefix}, each inserting lightweight trainable modules while freezing the backbone. LoRA is the mechanism used by \textsc{Any2Any}; Adapter and Prefix-Tuning serve as PEFT baselines. 

\subsubsection{Evaluation Metrics}
We evaluate from two complementary perspectives. During training, we monitor the core WBT reward (the \textit{tracking joint position} or \textit{tracking body position} reward) and compare methods by both converged reward and convergence speed, which reflect asymptotic tracking quality and sample efficiency. All such comparisons use the same compute budget and number of iterations.

During deployment in MuJoCo, all metrics are reported on the held-out OMOMO and LAFAN benchmarks. The \emph{success rate} is the fraction of evaluation clips tracked to completion without early termination; following the training termination criteria, a rollout fails once the root height error exceeds $0.25$\,m, the root orientation error exceeds $0.8$ (squared quaternion error, $\approx$$46^{\circ}$ tilt), or any tracked body-link height error exceeds $0.3$\,m. We further report the Mean Per-Joint Position Error (MPJPE, mm) for joint-level fidelity, the world-frame base position (cm) and orientation (deg) errors for global accuracy, and the mean per-frame velocity (mm/frame) magnitudes of the policy output as smoothness indicator (lower is smoother, a proxy for real-world deployability rather than tracking accuracy).

\subsection{Any2Any Transfer Performance }
% 在这个Result中
% To address \textbf{Q1}, we conduct two transfer experiments that together substantiate the \textbf{Any2Any} claim of our framework. In the first experiment, our Transformer-based WBT policy pretrained on LimX Oli 
% is transferred to three structurally distinct humanoids, Unitree G1, Unitree H1, 
% and LimX Luna. This demonstrates the robustness of \textsc{Any2Any} to target 
% robots with diverse morphologies. In the second experiment, Sonic, a strong open-source WBT policy with an VQ-VAE architecture originally trained on Unitree G1, is transferred to LimX Oli and Luna. This demonstrates the robustness of \textsc{Any2Any} to pretrained policies with different network structures.
% In both settings, \textsc{Any2Any} is compared against the Scratch baseline on the target embodiment.

% To address \textbf{Q1}, we evaluate \textsc{Any2Any} on our five transfer
% instances, each compared against the Scratch baseline on the target embodiment.
% We present the results by source backbone, starting with the external Sonic
% source and then our Oli-WBT source.

To answer \textbf{Q1}, we evaluate \textsc{Any2Any} on six cross-embodiment transfer instances and compare it with the training-from-scratch baseline under the same training budget. We analyze the results from two complementary perspectives: training time optimization, which measures convergence speed and converged reward, and held-out deployment performance, which measures whether the learned policy generalizes to unseen motions rather than merely optimizing the training reward.

% 为了进一步验证本方法的通用性，我们在目前开源的sota Sonic上做了实验，将原本g1上tracking policy迁移到Oli上。考虑到仅为了进行验证训练，我们仅使用了其中的Robot motion Encoder，FSQ，和g1 dyn decoder进行LoRA tuning。训练的奖励如图所示，可以看到，相比起从0训练的，LoRA的结果更快，且在未调整奖励参数的情况下甚至达到了更高的收敛的奖励，这体现了本文提出方法的巨大潜力。
\begin{figure}
    \centering
    \includegraphics[width=\linewidth]{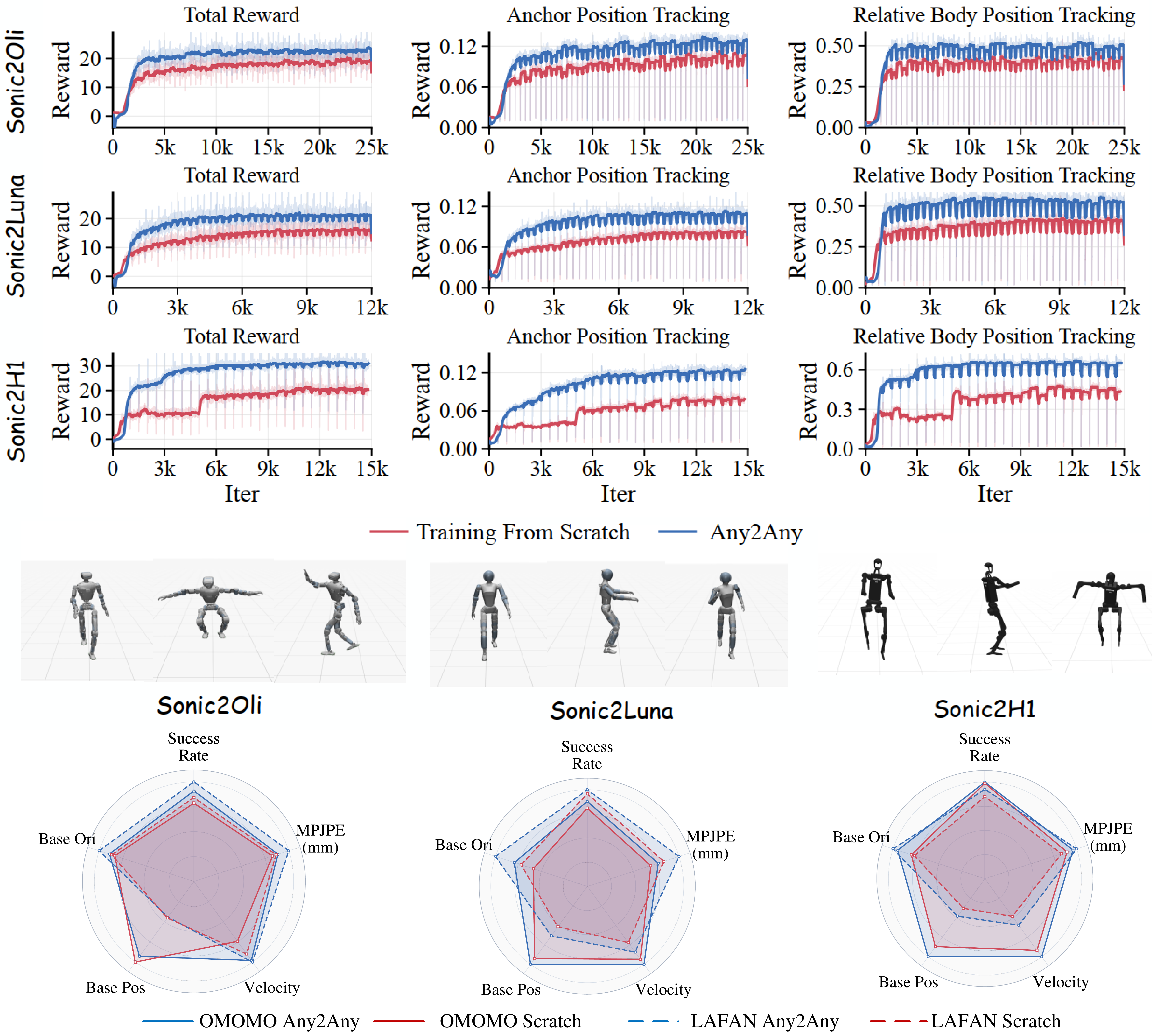}
    \caption{\textsc{Any2Any} transfer from Sonic to LimX humanoids, including
\textsc{Sonic2Oli} and \textsc{Sonic2Luna}. The curves compare \textsc{Any2Any}
with the baseline, and the snapshots show stable rollout motions
after adaptation.}
    \label{fig:sonic_transfer}
\end{figure}

\textbf{Sonic as source.}
For the external Sonic source (\textsc{Sonic2Oli}, \textsc{Sonic2Luna}, \textsc{Sonic2H1}), LoRA
modules are inserted into the actor dynamics decoder and the critic network,
while the FSQ module and other pretrained components remain frozen.

% 训得更好的，三个奖励更好。
%  对于三个代表性的拓扑，训练结果都是很好的
% 复用先验

As shown in Figure~\ref{fig:sonic_transfer}, \textsc{Any2Any} consistently transfers the external Sonic WBT prior to three representative target humanoids. These targets cover different types of embodiment gaps: \textsc{Sonic2Oli} involves transferring from the Unitree G1 source to a larger Oli humanoid with an inclined hip structure; \textsc{Sonic2Luna} further introduces parallel/closed-chain mechanisms; and \textsc{Sonic2H1} transfers to a lower-DoF serial-chain humanoid with a different body scale. Despite these different kinematic and dynamic gaps, \textsc{Any2Any} shows the same trend across all three targets: the total reward rises faster and converges to a higher value than Scratch, and the gains are also consistent in both anchor-position tracking and relative-body-position tracking rewards.

These results directly support the core design of \textsc{Any2Any}. Training from scratch must rediscover whole-body tracking, balance regulation, and embodiment-specific coordination entirely from target-side interaction. In contrast, \textsc{Any2Any} first uses kinematic alignment to place the target observations and actions into the source policy's semantic space, so that the pretrained Sonic WBT prior can be reused. LoRA then adapts only the remaining target-specific dynamics residual. Therefore, the faster convergence and higher converged rewards across Oli, Luna, and H1 show that \textsc{Any2Any} provides an efficient post-training paradigm for reusing robot-specific WBT policies across different humanoid embodiments.

% As shown in Figure~\ref{fig:sonic_transfer}, across all three targets, \textsc{Any2Any} converges faster and reaches higher rewards than training from scratch.
% On \textsc{Sonic2Oli}, \textsc{Any2Any} rapidly increases the total reward within the first few thousand iterations and converges to roughly 22, while the Scratch baseline remains around 18.
% Similar trends are observed for the anchor-position and relative-body-position rewards, where \textsc{Any2Any} consistently achieves higher values, indicating better global body tracking and local whole-body configuration tracking.
% The largest gain is observed on \textsc{Sonic2H1}. In fact, both G1 and H1 belong to the Unitree serial-chain family and do not require inclined-hip or closed-chain correction (see Table 1).
% Instead, H1 introduces a strong reduction in controllable DoF and a clear body-scale change compared with the Sonic source robot G1. 
% Training from scratch must therefore discover whole-body tracking, balance, and reduced DoF compensation directly on the target embodiment, which leads to slower and less stable optimization. 
% In contrast, \textsc{Any2Any} can still reuse the semantically matched G1 WBT prior through kinematic alignment, while LoRA only needs to adapt the remaining target-specific dynamics.
% This explains why the improvement is particularly pronounced on \textsc{Sonic2H1}.

The radar plots further evaluate whether the training-time gains transfer to unseen motion distributions. We report results on two held-out benchmarks with complementary properties: OMOMO and LAFAN.
For \textsc{Sonic2Oli}, \textsc{Any2Any} consistently improves over Scratch on both benchmarks.
On LAFAN, the improvement is even more pronounced: the success rate increases from 83.1 $\%$ to 90.6 $\%$ , MPJPE decreases from 69.7 mm to 47.2 mm, and the base-position and base-orientation errors decrease from 10.48 cm to 8.49 cm and from 7.3$^\circ$ to 5.6$^\circ$.
These results indicate that the transferred Sonic prior improves both manipulation-style tracking and dynamic locomotion tracking on Oli.
For \textsc{Sonic2Luna}, \textsc{Any2Any} also provides consistent gains despite the larger structural mismatch introduced by Luna's inclined hip and closed-chain/parallel mechanisms.
For \textsc{Sonic2H1}, \textsc{Any2Any} mainly improves stability, joint-level fidelity, orientation tracking, and smoothness.
Overall, the held-out radar results show that \textsc{Any2Any} does not merely optimize the training reward. Across OMOMO and LAFAN, it generally improves completion rate, joint-level tracking, root-orientation tracking, and action smoothness.

\begin{figure}
    \centering
\includegraphics[width=\linewidth]{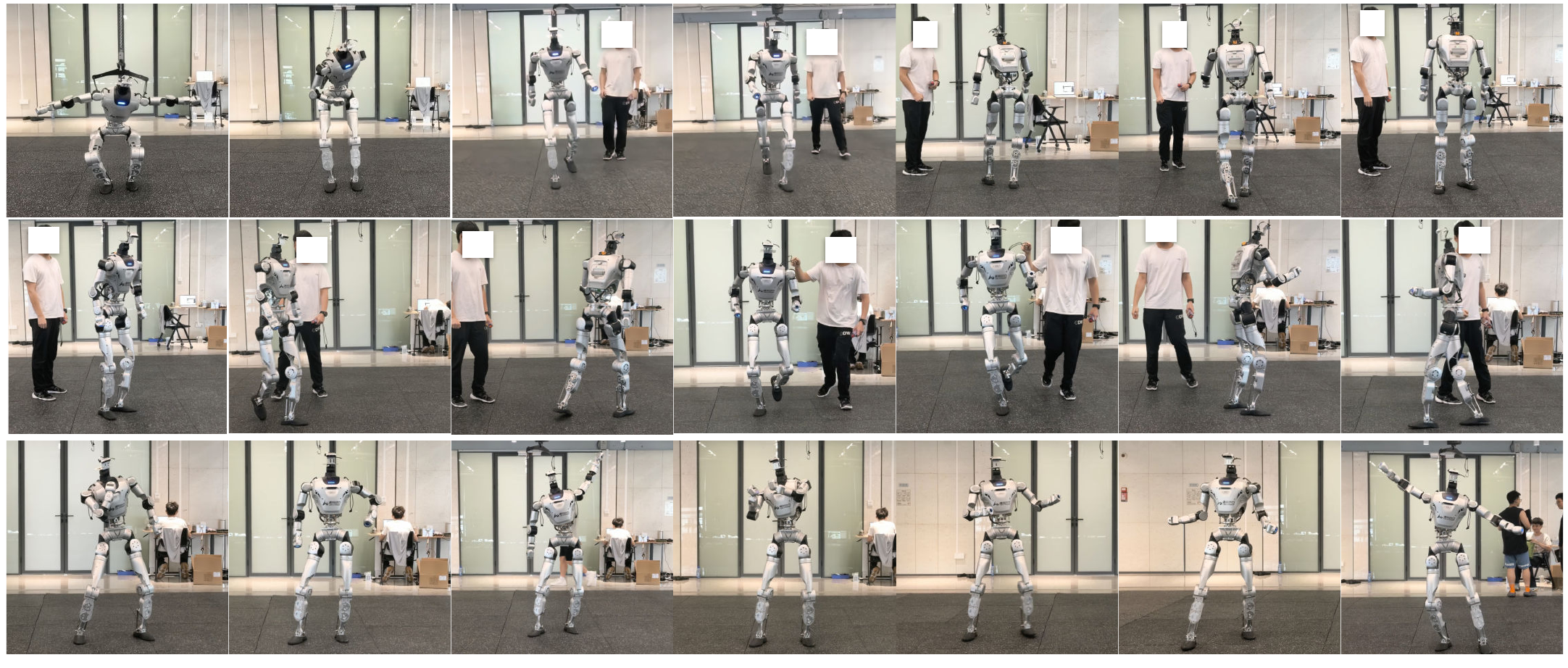}
    \caption{ Snapshots of the \textsc{Sonic2Oli} real-world experiments.  Trained on the AMASS dataset for 90 GPU-hours, the policy performs a variety of behaviors on the LimX Oli, including walking, running, turning, and dancing.
 }
    \label{fig:realsnapshot}
\end{figure}

The rollout snapshots provide qualitative evidence that the adapted policies remain stable across different target humanoids and diverse motions. Together with the faster reward convergence and improved held-out tracking metrics, these results indicate that \textsc{Any2Any} does not learn whole-body control from scratch. Instead, it preserves the reusable WBT prior from the source policy through kinematic alignment and only learns a compact target-specific dynamics correction. Beyond simulation, we further deploy the adapted \textsc{Sonic2Oli} policy on the physical LimX Oli, where it reliably reproduces a range of generalized behaviors such as walking, running, turning, and dancing (Figure~\ref{fig:realsnapshot}), confirming that the transferred prior remains effective under real-world dynamics.

\begin{figure}[h]
    \centering
    \includegraphics[width=1.0\linewidth]{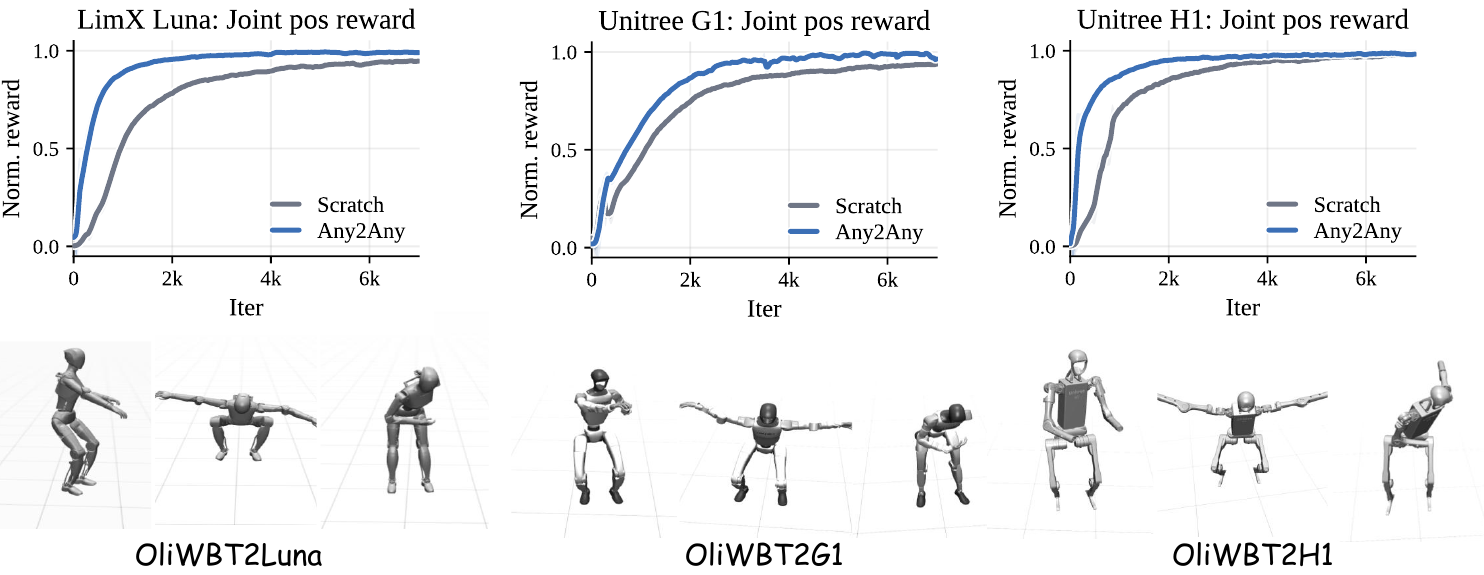}
    \caption{\textsc{Any2Any} transfer from Oli-WBT to three target
humanoids: \textsc{OliWBT2Luna}, \textsc{OliWBT2G1}, and
\textsc{OliWBT2H1}. \textsc{Any2Any} is compared with the baseline
trained from scratch. (a) Tracking-error radar plots.
(b) Training curves of normalized tracking reward.
(c) Sim-to-sim rollouts on diverse motions.}
    \label{fig:main_transfer}
\end{figure}

\textbf{Oli-WBT as source.}
We further evaluate whether the same transfer paradigm also holds for our self-pretrained Oli-WBT source policy. As shown in Figure~\ref{fig:main_transfer}, \textsc{Any2Any} transfers Oli-WBT to three representative target humanoids: \textsc{OliWBT2Luna}, \textsc{OliWBT2G1}, and \textsc{OliWBT2H1}.

The training curves show a consistent optimization advantage. On all three targets, \textsc{Any2Any} reaches the high-reward regime much earlier than Scratch. For \textsc{OliWBT2Luna}, \textsc{Any2Any} rapidly increases the normalized joint-position reward.
The same trend appears on \textsc{OliWBT2G1} and \textsc{OliWBT2H1}, \textsc{Any2Any} enters the high-reward region earlier and converges to a higher or comparable final reward.
This indicates that the pretrained Oli-WBT policy already contains reusable joint-level tracking and whole-body coordination priors, and the target policy does not need to rediscover them from scratch.

The radar plots further show that the training-time gains transfer to unseen motion distributions. On OMOMO, which emphasizes whole-body object-manipulation motions, \textsc{Any2Any} generally improves joint-level and root tracking.
On LAFAN, which contains more dynamic locomotion motions, \textsc{Any2Any} also maintains clear advantages in stability and tracking accuracy.
These results indicate that the transferred Oli-WBT prior generalizes to both manipulation-style and high-dynamic locomotion motions.

Overall, the Oli-WBT results are consistent with the Sonic-transfer results and further support the core hypothesis of \textsc{Any2Any}. Across Luna, G1, and H1, \textsc{Any2Any} achieves faster reward convergence and stronger held-out tracking performance than training from scratch. This shows that the proposed post-training paradigm is not tied to a specific pretrained backbone. By first aligning the target robot with the source policy's kinematic semantic space and then adapting only a compact dynamics residual, \textsc{Any2Any} efficiently reuses robot-specific WBT priors across different humanoid embodiments.

\subsection{Data and Compute Efficiency}
\label{sec:ablation_efficiency}
\begin{figure}
    \centering
    \includegraphics[width=\linewidth]{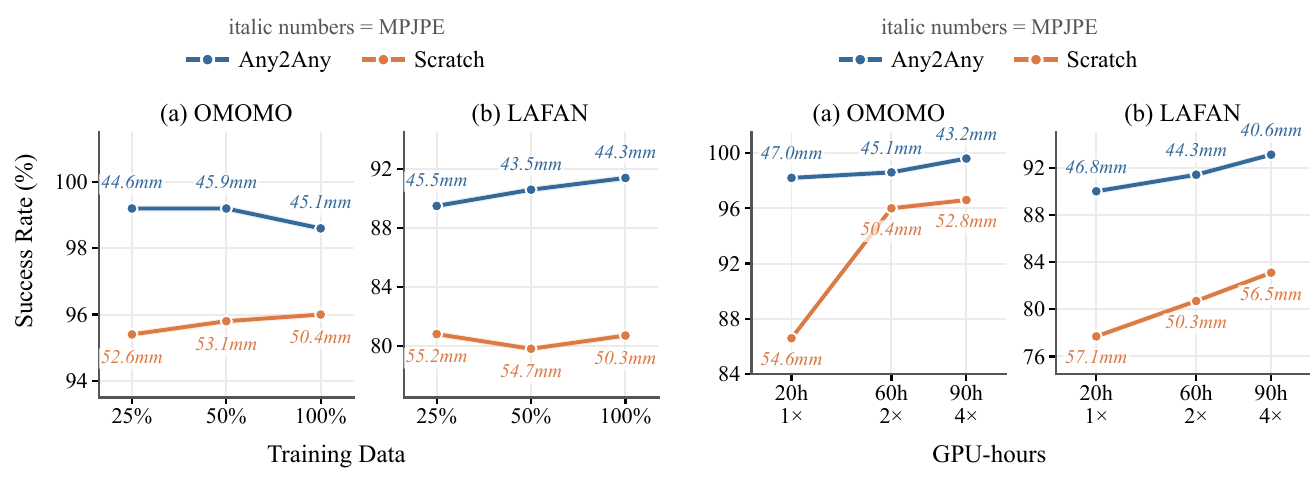}
    \caption{\textbf{Data and compute efficiency on \textsc{Sonic2Oli}.}
(a)~Held-out success rate (MPJPE annotated) on OMOMO and LAFAN as the adaptation
data is scaled over $25/50/100\%$ of AMASS under a fixed iteration budget.
(b)~Training reward versus parallel-sampling budget (one/two/four A100 GPUs at
$8{,}000$ iterations). \textsc{Any2Any} stays above training-from-scratch across
all data and compute budgets.}
    \label{fig:data_compute}
\end{figure}

To address \textbf{Q2}, we study how strongly \textsc{Any2Any} depends on the
amount of adaptation data and on the parallel-simulation budget. All experiments
in this section use the unified \textsc{Sonic2Oli} task and compare
\textsc{Any2Any} against training from scratch under matched settings
(Figure~\ref{fig:data_compute}).

\textbf{Data efficiency.} We vary the amount of adaptation data, using $25\%$,
$50\%$, and $100\%$ of the AMASS corpus, train every variant for the same number
of iterations, and evaluate on both held-out benchmarks. On both OMOMO and LAFAN,
\textsc{Any2Any} achieves a higher success rate and better joint-level tracking
than scratch at every data scale. Interestingly, on OMOMO the success rate and
tracking accuracy of \textsc{Any2Any} slightly \emph{decrease} as more data is
added. We attribute this to the near-static nature of OMOMO manipulation motions:
a small amount of in-place data is already sufficient to learn the relatively
simple quasi-static tracking, so additional data yields little benefit and mostly
adds variance. On LAFAN, whose motions are highly dynamic, the expected trend
emerges: \textsc{Any2Any} improves markedly as more data becomes available,
whereas scratch stays at a consistently low level. Overall, the transferred WBT prior lets the target policy reach broad whole-body
tracking from far less data than training from scratch.

\textbf{Compute efficiency.} We then fix the data and vary the parallel-sampling
budget, using one, two, and four A100 GPUs with the same number of environments
per GPU, and compare all runs at $8{,}000$ iterations. As the GPU budget grows,
\textsc{Any2Any} improves only slightly, while scratch fluctuates substantially
yet stays below \textsc{Any2Any} throughout. That is, scratch relies heavily on
large-scale parallel sampling to discover whole-body control, whereas
\textsc{Any2Any} already reaches a near-complete result under a small compute
budget.

Together, these two studies show that, by reusing the WBT motion prior,
\textsc{Any2Any} is far less dependent on large adaptation datasets and
large-scale parallel simulation than training from scratch, attaining strong
tracking under limited data and compute.

% 为了说明Any2Any对数据和计算效率没那么依赖，我们在Sonic2Oli这一统一任务下进行了不同数据量和不同GPU hours的对比实验。在数据量实验方面，我们分别选择百分之25，百分之50和百分之100的AMASS数据进行训练相同的迭代次数，并均在OMOMO和LAFAN上进行比较，我们发现两个数据集下，Any2Any都比从0训练的的成功率更高，且局部追踪更好。值得注意的是OMOMO数据中，随着数据量的增加，成功率和追踪出现了轻微的下滑，我们认为是因为少量的原地数据即可让其学到相对简单的相对静止追踪，而LAFAN这种高动态动作的变化就符合我们的印象，在Any2Any中，随着数据量提高出现了显著的增加，而Scratch的训练中，始终维持在一个较低的水平。
% GPU hours的实验中，我们分别选择单卡A100,双卡A100和四卡A100，每张卡上相同的环境数量，并都在8000iter进行对比，我们发现，随着GPU hours的增加，Any2Any略有增加，而Scratch出现了明显的波动，但始终低于Any2Any。这两个结果说明结合了WBT的运动先验，是可以在小量GPU下实现一个相对完整的结果的。

\subsection{Ablation Analysis}
\label{sec:ablation_arch}

To address \textbf{Q3}, we conduct architectural ablations on the representative
\textsc{OliWBT2Luna} transfer instance using both Transformer and MLP backbones.
These ablations examine whether the two stages of \textsc{Any2Any} are both
necessary: kinematic alignment for resolving structural mismatch, and localized
LoRA adaptation for absorbing the remaining dynamics residual.

% To answer Q3,
% We conduct architectural ablations on the representative
% \textsc{OliWBT2Luna} transfer instance to analyze the two core components of
% \textsc{Any2Any}: kinematic alignment and parameter-efficient adaptation, using both Transformer and MLP backbones.

% we ablate the key architectural components of \textsc{Any2Any},
% including kinematic alignment and the choice of parameter-efficient adaptation
% mechanism. All experiments in this section are conducted under the same
% Oli-to-Luna transfer setting, using both Transformer and MLP backbones.

\begin{figure}[t]
\centering

% ================= TABLE: FULL WIDTH =================
\footnotesize
\setlength{\tabcolsep}{3.2pt}
\renewcommand{\arraystretch}{1.08}

\begin{tabular*}{\linewidth}{@{\extracolsep{\fill}}lcccccc@{}}
\toprule
Method
& \makecell{Trainable\\Params$\downarrow$}
& \makecell{FPS$\uparrow$}
& \makecell{Collection\\Cost$\downarrow$}
& \makecell{Learning\\Cost$\downarrow$}
& \makecell{Joint\\Reward$\uparrow$}
& \makecell{Total\\Reward$\uparrow$} \\
\midrule
Full FT (w/ align)
& 100.00\% 
& 173.0 k 
& \textbf{3.90}
& 4.48
& 0.43  
& 19.04 \\
\textsc{Any2Any} (LoRA) 
& \textbf{5.26\%} 
& \textbf{196.0 k} 
&  4.53
& \textbf{3.07} 
& \textbf{0.54}
&  \textbf{23.86}\\
\bottomrule
\end{tabular*}

\vspace{0.6em}

% ================= FIGURE =================
\includegraphics[width=\linewidth]
{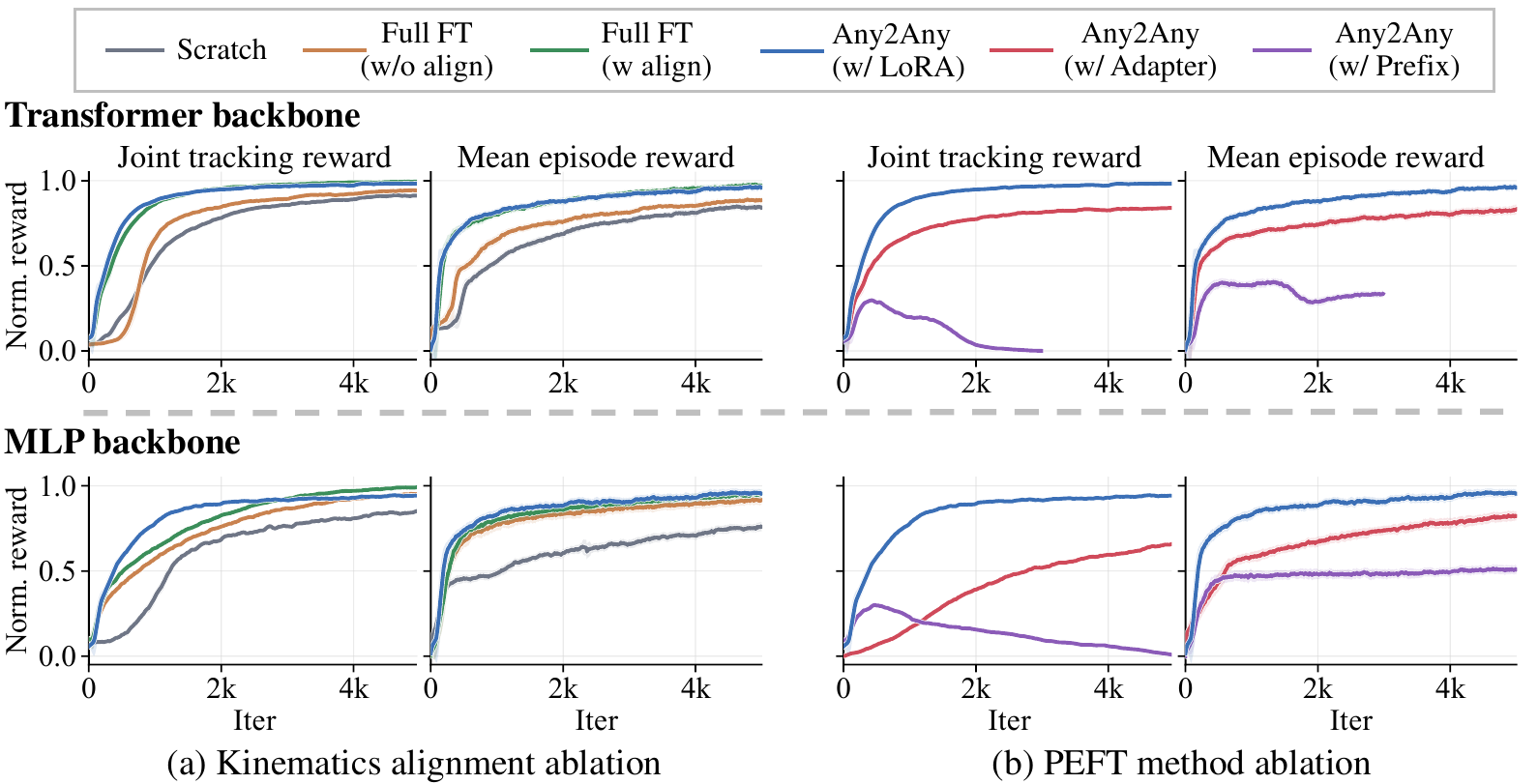}

\caption{
Ablation of \textsc{Any2Any} architectural components on
\textsc{OliWBT2Luna}. The top table compares aligned full fine-tuning and
\textsc{Any2Any} with LoRA. (a) Kinematic alignment ablation. (b) PEFT method
ablation under kinematic alignment. \textsc{Any2Any}-LoRA achieves comparable
rewards to full fine-tuning while using fewer trainable parameters and lower
training cost.
}
\label{fig:ablation}

\end{figure}

Figure~\ref{fig:ablation}(a) studies the effect of kinematic alignment. We
compare four settings: training from scratch, full fine-tuning without
alignment, full fine-tuning with alignment, and \textsc{Any2Any} with LoRA.
Across both Transformer and MLP backbones, training from scratch converges
slowly and reaches a lower final reward, indicating that learning whole-body
tracking directly on the target embodiment remains data- and compute-intensive.
Full fine-tuning without alignment improves over scratch, but its convergence
and final reward are still limited. This suggests that the pretrained WBT prior
cannot be effectively reused when the target robot's observations and actions
are interpreted under inconsistent joint semantics. After kinematic alignment is
introduced, full fine-tuning converges faster and achieves higher tracking
rewards, confirming that aligning the observation, reference, and action spaces
is necessary before transferring a pretrained policy across embodiments.

Notably, \textsc{Any2Any} with LoRA achieves performance comparable to aligned
full fine-tuning while updating only a small fraction of the policy parameters.
As shown in the table above Figure~\ref{fig:ablation}, LoRA reduces the
trainable parameters from 100\% to 5.26\%, improves FPS from 34.8 k to 111.7 k.
Although its collection cost is slightly higher, the overall adaptation cost is lower, and the final joint and total rewards increase from 0.43 and 19.04 to 0.54 and 23.86.
These results show that, once kinematic
semantics are aligned, adapting the entire pretrained policy is unnecessary:
a lightweight low-rank update is sufficient to specialize the source WBT prior
to the target embodiment.

Figure~\ref{fig:ablation}(b) further compares different PEFT mechanisms under
the same aligned setting. LoRA consistently provides the best balance between
training stability, convergence speed, and final performance for both
Transformer and MLP backbones. Adapter tuning also improves over training from
scratch, but converges more slowly and reaches lower rewards, suggesting that
its additional bottleneck modules are harder to optimize in closed-loop
whole-body control. Prefix tuning is the least stable: its reward either
saturates at a low value or degrades during training, indicating that modifying
only the input conditioning is insufficient to compensate for embodiment-level
dynamical mismatch. Overall, these results support the design choice of
\textsc{Any2Any}: kinematic alignment first establishes a shared semantic space,
and LoRA then provides an efficient residual adaptation mechanism for the
remaining embodiment-specific dynamics.

\begin{wrapfigure}{r}{0.6\textwidth}
\vspace{-1.2em}
\centering

% ================= TABLE =================
\scriptsize
\setlength{\tabcolsep}{2.3pt}
\renewcommand{\arraystretch}{1.05}

\begin{tabular}{lccccccc}
\toprule
\textbf{Setting}
& \multicolumn{4}{c}{\textbf{Actor}}
& \multicolumn{3}{c}{\textbf{Critic}} \\
\cmidrule(lr){2-5}\cmidrule(lr){6-8}
& \textbf{Backbone}
& \textbf{Ref. In.}
& \textbf{Prop. In.}
& \textbf{Out.}
& \textbf{Backbone}
& \textbf{In.}
& \textbf{Out.} \\
\midrule
S1  & $\checkmark$ &              &              &              &              &              &              \\
S2  & $\checkmark$ &              &              &              & $\checkmark$ &              &              \\
S3  & $\checkmark$ & $\checkmark$ &              &              & $\checkmark$ &              &              \\
S4  & $\checkmark$ &              & $\checkmark$ &              & $\checkmark$ &              &              \\
S5  & $\checkmark$ &              &              & $\checkmark$ & $\checkmark$ &              &              \\
S6  & $\checkmark$ &              &              &              & $\checkmark$ & $\checkmark$ &              \\
\anytxt{\textbf{S7}}
    & \anycheck
    & 
    & \anycheck
    & \anycheck
    & \anycheck
    & 
    &  \\
S8  & $\checkmark$ & $\checkmark$ & $\checkmark$ & $\checkmark$ &              &              &              \\
S9  & $\checkmark$ & $\checkmark$ & $\checkmark$ & $\checkmark$ & $\checkmark$ & $\checkmark$ & $\checkmark$ \\
\bottomrule
\end{tabular}
\vspace{0.6em}

% S8  &              & $\checkmark$ & $\checkmark$ & $\checkmark$ &              &              &              \\

% ================= FIGURE =================
\includegraphics[width=\linewidth]
{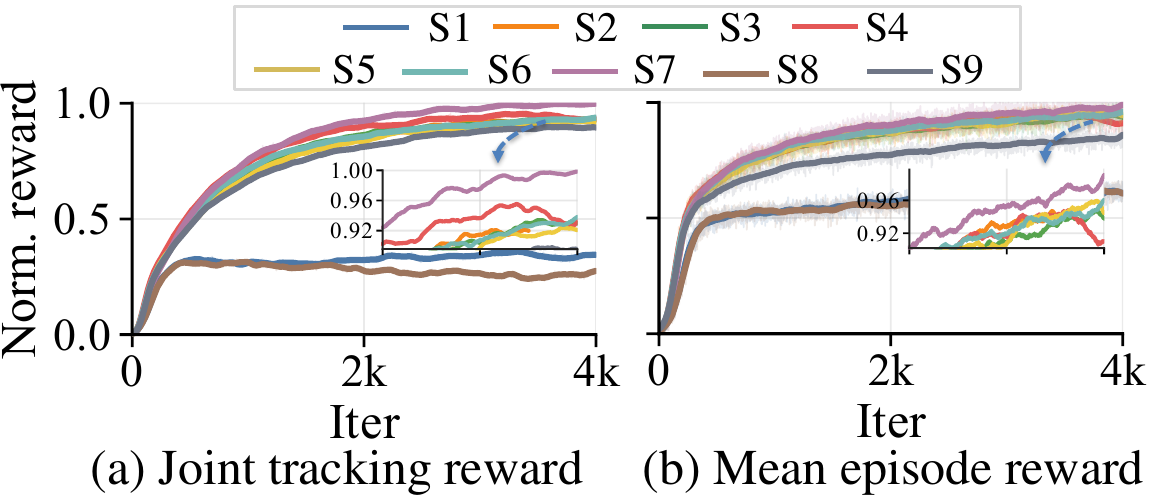}

\caption{
Ablation of LoRA injection scopes on \textsc{OliWBT2Luna}. The table summarizes
the component-level injection locations across actor and critic modules, while
the curves show the resulting joint tracking reward and mean episode reward.
}
\label{fig:lora_scope}

\vspace{-1.2em}
\end{wrapfigure}

% \begin{wrapfigure}{r}{0.6\textwidth}
% \vspace{-1.2em}
% \centering

% % ================= TABLE =================
% \scriptsize
% \setlength{\tabcolsep}{3pt}
% \renewcommand{\arraystretch}{1.08}

% \begin{tabularx}{\linewidth}{lcc>{\raggedright\arraybackslash}X}
% \toprule
% \textbf{Setting} 
% & \textbf{Scratch} 
% & \textbf{Backbone-LoRA} 
% & \textbf{Extra LoRA Injection} \\
% \midrule
% Specialist 
% & $\checkmark$ 
% & 
% & -- \\

% S1 
% & 
% & $\checkmark$ 
% & -- \\

% S2 
% & 
% & $\checkmark$ 
% & Actor/Critic In--Out Proj. \\

% S3 
% & 
% & $\checkmark$ 
% & Actor Input: Ref. Proj. \\

% S4
% & 
% & $\checkmark$ 
% & Actor Input: Prop. + Act. Proj. \\
% \bottomrule
% \end{tabularx}

% \vspace{0.6em}

% % ================= FIGURE =================
% \includegraphics[width=\linewidth]
% {corl_2026_template_submission/figure/different_scopes_reward.pdf}

% \caption{
% Performance comparison under different LoRA adaptation scopes. The table
% summarizes the component-level injection locations, while the curves show the
% training performance of each scope.
% }
% \label{fig:lora_scope}

% \vspace{-1.2em}
% \end{wrapfigure}

Figure~\ref{fig:lora_scope} further studies where the LoRA residual should be
injected after kinematic alignment. The table groups the candidate injection
sites into actor and critic components, including the actor backbone, reference
input projection, proprioception input projection, action output projection,
critic backbone, and critic input/output projections. This component-level
ablation allows us to examine which parts of the pretrained WBT policy carry
embodiment-agnostic motion priors and which parts require target-specific
dynamics correction.

% The results show that the injection location has a significant effect on both
% convergence and final tracking performance. Applying LoRA only to the actor
% backbone provides a useful baseline adaptation, confirming that the frozen
% pretrained policy contains reusable whole-body motion priors. However, adapting
% the backbone alone is not sufficient to fully compensate for the target
% embodiment. 

The best performance is achieved by S7, which applies LoRA to the
actor backbone, actor proprioception input projection, actor output projection,
and critic backbone. This indicates that the dominant residual mismatch after
kinematic alignment lies in the dynamics-aware control pathway: the policy needs
to reinterpret the target robot's proprioceptive state and recent action
history, and also adjust how the shared motion representation is decoded into
target-specific joint commands.

This finding is consistent with the design principle of \textsc{Any2Any}. After
the observation, reference, and action spaces are kinematically aligned, the
reference-motion stream becomes relatively embodiment-agnostic and should be
largely preserved. 
In contrast, the
proprioception stream, action output head, and critic backbone are more directly
coupled to the target robot's mass distribution, actuator response, contact
behavior, and closed-loop stability. Adapting these modules allows the policy to
absorb the remaining dynamics residual while keeping the high-level WBT prior
intact.
Therefore, S7 provides the best trade-off between target-specific dynamics
adaptation and preservation of reusable whole-body motion priors.

Overall, the ablations validate the design of \textsc{Any2Any}. Kinematic alignment is necessary to establish a shared semantic space; LoRA is more effective than alternative PEFT mechanisms for closed-loop humanoid WBT adaptation; and the best injection scope is concentrated on dynamics-sensitive modules rather than the entire policy. These results support the central decomposition of \textsc{Any2Any}: structural mismatch is handled by alignment, while the remaining dynamics gap is handled by localized low-rank adaptation.

\section{Discussion and Conclusion}

We presented \textsc{Any2Any}, a post-training paradigm that transfers a pretrained robot-specific whole-body tracking (WBT) policy to a new humanoid using only about 1\% of the data and compute of training from scratch. \textsc{Any2Any} rests on a single decomposition of the cross-embodiment gap: a structural \emph{kinematic} mismatch, resolved by an alignment that renders the target robot's observations and actions interpretable to the frozen source policy, and a residual \emph{dynamics} gap, absorbed by a compact low-rank correction injected only into the dynamics-sensitive modules. Across six transfers spanning two pretrained backbones (Oli-WBT and Sonic) and four humanoids (LimX Oli, LimX Luna, Unitree G1, and H1), this recipe yields faster convergence, stronger held-out tracking, and reliable real-world deployment, while modifying only about 5\% of the policy parameters.

We read this result as more than an efficient fine-tuning trick. To the best of our knowledge, it is the first evidence that a large-scale WBT prior is \emph{not} intrinsically bound to the embodiment on which it was trained. We attribute this to a deeper structure: despite their diverse morphologies, humanoids are implicitly anchored to a common, human-centric motor manifold---their motion data is sourced from humans, their movements imitate human behavior, and their mechanical design itself references the human body. A pretrained WBT policy therefore encodes reusable structure such as balance regulation, inter-limb coordination, motion timing, and contact-aware tracking, which is largely shared across embodiments rather than specific to one robot. Once the embodiment-specific observation--action mismatch is removed, much of this prior can be preserved and reused, leaving only a low-dimensional, embodiment-specific dynamics residual to be learned. This is precisely why such aggressive data and compute savings are attainable, and we regard it as an encouraging signal that future humanoid policies can be explored far more boldly than the per-robot, train-from-scratch convention suggests.

This view also suggests a concrete design principle for cross-embodiment WBT: treat a compute-heavy WBT backbone as a reusable building block for motion understanding and whole-body coordination, and equip each new humanoid with only a lightweight kinematic mapping and a small dynamics adapter. Rather than retraining a controller for every robot, or rebuilding a multi-robot generalist from scratch, one starts from an existing expert and pays only a small adaptation cost, accelerating deployment on new platforms.

Looking forward, our findings point to two complementary directions for making cross-embodiment transfer the rule rather than the exception. \emph{First}, the breadth of our experiments leads us to advocate for a community \emph{Reference Humanoid}: a standardized platform designed to be simultaneously human-like and easy to transfer to, serving as a common hub and benchmark for cross-embodiment transfer, and accelerating research not only on basic whole-body motion but also on higher-level skills such as loco-manipulation and dexterous manipulation. \emph{Second}, and complementarily, transferability can be built in at pretraining time: training across a diverse set of robots together with structural randomization should yield priors that are inherently easier to migrate to unseen embodiments. These two directions reinforce each other: a shared reference platform gives multi-robot pretraining a natural anchor, while priors engineered for transfer make each new humanoid cheaper to bring online.

In summary, for the first time, we show that a large-scale humanoid WBT policy can be transferred to a new humanoid with only a small fraction of the original data and compute. \textsc{Any2Any} is only the beginning of this journey, but it suggests that reusable, cross-embodiment whole-body intelligence is within reach.

\clearpage
% The acknowledgments are automatically included only in the final and preprint versions of the paper.
\acknowledgments{}

%===============================================================================

\bibliography{example}  % .bib

@article{yuan2025survey,
  title={A survey of behavior foundation model: Next-generation whole-body control system of humanoid robots},
  author={Yuan, Mingqi and Yu, Tao and Ge, Wenqi and Yao, Xiuyong and Li, Dapeng and Wang, Huijiang and Chen, Jiayu and Li, Bo and Zhang, Wei and Zeng, Wenjun and others},
  journal={IEEE transactions on pattern analysis and machine intelligence},
  year={2025},
  publisher={IEEE}
}

@article{cetin2024finer,
  title={Finer behavioral foundation models via auto-regressive features and advantage weighting},
  author={Cetin, Edoardo and Touati, Ahmed and Ollivier, Yann},
  journal={arXiv preprint arXiv:2412.04368},
  year={2024}
}

@article{cheng2024expressive,
  title={Expressive whole-body control for humanoid robots},
  author={Cheng, Xuxin and Ji, Yandong and Chen, Junming and Yang, Ruihan and Yang, Ge and Wang, Xiaolong},
  journal={arXiv preprint arXiv:2402.16796},
  year={2024}
}

@article{he2024omnih2o,
  title={Omnih2o: Universal and dexterous human-to-humanoid whole-body teleoperation and learning},
  author={He, Tairan and Luo, Zhengyi and He, Xialin and Xiao, Wenli and Zhang, Chong and Zhang, Weinan and Kitani, Kris and Liu, Changliu and Shi, Guanya},
  journal={arXiv preprint arXiv:2406.08858},
  year={2024}
}

@article{gu2026humanoid,
  title={Humanoid locomotion and manipulation: Current progress and challenges in control, planning, and learning},
  author={Gu, Zhaoyuan and Li, Junheng and Shen, Wenlan and Yu, Wenhao and Xie, Zhaoming and McCrory, Stephen and Cheng, Xianyi and Shamsah, Abdulaziz and Griffin, Robert and Liu, C Karen and others},
  journal={IEEE/ASME Transactions on Mechatronics},
  volume={31},
  number={2},
  pages={2300--2330},
  year={2026},
  publisher={IEEE}
}

@article{li2025bfm,
  title={Bfm-zero: A promptable behavioral foundation model for humanoid control using unsupervised reinforcement learning},
  author={Li, Yitang and Luo, Zhengyi and Zhang, Tonghe and Dai, Cunxi and Kanervisto, Anssi and Tirinzoni, Andrea and Weng, Haoyang and Kitani, Kris and Guzek, Mateusz and Touati, Ahmed and others},
  journal={arXiv preprint arXiv:2511.04131},
  year={2025}
}

@inproceedings{pirotta2024fast,
  title={Fast imitation via behavior foundation models},
  author={Pirotta, Matteo and Tirinzoni, Andrea and Touati, Ahmed and Lazaric, Alessandro and Ollivier, Yann},
  booktitle={International Conference on Learning Representations},
  volume={2024},
  pages={12685--12724},
  year={2024}
}

@article{peng2018deepmimic,
  title={DeepMimic: Example-Guided Deep Reinforcement Learning of Physics-Based Character Skills},
  author={Peng, Xue Bin and Abbeel, Pieter and Levine, Sergey and van de Panne, Michiel},
  journal={ACM Transactions on Graphics},
  volume={37},
  number={4},
  pages={1--14},
  year={2018},
  doi={10.1145/3197517.3201311},
  url={https://arxiv.org/abs/1804.02717}
}

@article{ze2025twist,
  title={Twist: Teleoperated whole-body imitation system},
  author={Ze, Yanjie and Chen, Zixuan and Ara{\'u}jo, Joao Pedro and Cao, Zi-ang and Peng, Xue Bin and Wu, Jiajun and Liu, C Karen},
  journal={arXiv preprint arXiv:2505.02833},
  year={2025}
}

@article{chen2025gmt,
  title={Gmt: General motion tracking for humanoid whole-body control},
  author={Chen, Zixuan and Ji, Mazeyu and Cheng, Xuxin and Peng, Xuanbin and Peng, Xue Bin and Wang, Xiaolong},
  journal={arXiv preprint arXiv:2506.14770},
  year={2025}
}

@article{luo2025sonic,
  title={Sonic: Supersizing motion tracking for natural humanoid whole-body control},
  author={Luo, Zhengyi and Yuan, Ye and Wang, Tingwu and Li, Chenran and Chen, Sirui and Castaneda, Fernando and Cao, Zi-Ang and Li, Jiefeng and Minor, David and Ben, Qingwei and others},
  journal={arXiv preprint arXiv:2511.07820},
  year={2025}
}

@inproceedings{li2025clone,
  title={CLONE: Closed-Loop Whole-Body Humanoid Teleoperation for Long-Horizon Tasks},
  author={Li, Yixuan and Lin, Yutang and Cui, Jieming and Liu, Tengyu and Liang, Wei and Zhu, Yixin and Huang, Siyuan},
  booktitle={Proceedings of The 9th Conference on Robot Learning},
  year={2025},
  url={https://arxiv.org/abs/2506.08931}
}

@article{xue2026xhugwbc,
  title={Scalable and General Whole-Body Control for Cross-Humanoid Locomotion},
  author={Xue, Yufei and Lin, YunFeng and Dong, Wentao and Tang, Yang and Wang, Jingbo and Pang, Jiangmiao and Zhou, Ming and Liu, Minghuan and Zhang, Weinan},
  journal={arXiv preprint arXiv:2602.05791},
  year={2026}
}

@article{gupta2022metamorph,
  title={Metamorph: Learning universal controllers with transformers},
  author={Gupta, Agrim and Fan, Linxi and Ganguli, Surya and Fei-Fei, Li},
  journal={arXiv preprint arXiv:2203.11931},
  year={2022}
}

@inproceedings{liu2025locoformer,
  title={LocoFormer: Generalist Locomotion via Long-context Adaptation},
  author={Liu, Min and Pathak, Deepak and Agarwal, Ananye},
  booktitle={Proceedings of The 9th Conference on Robot Learning},
  year={2025}
}

@article{yang2025multiloco,
  title={Multi-Loco: Unifying Multi-Embodiment Legged Locomotion via Reinforcement Learning Augmented Diffusion},
  author={Yang, Shunpeng and Fu, Zhen and Cao, Zhefeng and Guo, Junde and Wensing, Patrick and Zhang, Wei and Chen, Hua},
  journal={arXiv preprint arXiv:2506.11470},
  year={2025}
}

@article{lin2025hzero,
  title={H-Zero: Cross-Humanoid Locomotion Pretraining Enables Few-shot Novel Embodiment Transfer},
  author={Lin, Yunfeng and Liu, Minghuan and Xue, Yufei and Zhou, Ming and Yu, Yong and Pang, Jiangmiao and Zhang, Weinan},
  journal={arXiv preprint arXiv:2512.00971},
  year={2025},
  url={https://arxiv.org/abs/2512.00971}
}

@article{bjorck2025groot,
  title={GR00T N1: An Open Foundation Model for Generalist Humanoid Robots},
  author={Bjorck, Johan and Casta{\~n}eda, Fernando and Cherniadev, Nikita and Da, Xingye and Ding, Runyu and Fan, Linxi and Fang, Yu and Fox, Dieter and Hu, Fengyuan and Huang, Spencer and others},
  journal={arXiv preprint arXiv:2503.14734},
  year={2025}
}

@article{luo2026being,
  title={Being-H0.5: Scaling Human-Centric Robot Learning for Cross-Embodiment Generalization},
  author={Luo, Hao and Wang, Ye and Zhang, Wanpeng and Zheng, Sipeng and Xi, Ziheng and Xu, Chaoyi and Xu, Haiweng and Yuan, Haoqi and Zhang, Chi and Wang, Yiqing and others},
  journal={arXiv preprint arXiv:2601.12993},
  year={2026}
}

@article{bai2026hex,
  title={HEX: Humanoid-Aligned Experts for Cross-Embodiment Whole-Body Manipulation},
  author={Bai, Shuanghao and Li, Meng and Lv, Xinyuan and Wang, Jiawei and Wang, Xinhua and Liao, Fei and Hou, Chengkai and Gu, Langzhe and Zhou, Wanqi and Wu, Kun and others},
  journal={arXiv preprint arXiv:2604.07993},
  year={2026}
}

@article{ding2023peft,
  title={Parameter-efficient Fine-tuning of Large-scale Pre-trained Language Models},
  author={Ding, Ning and Qin, Yujia and Yang, Guang and Wei, Fuchao and Yang, Zonghan and Su, Yusheng and Hu, Shengding and Chen, Yulin and Chan, Chi-Min and Chen, Weize and others},
  journal={Nature Machine Intelligence},
  volume={5},
  number={3},
  pages={220--235},
  year={2023},
  doi={10.1038/s42256-023-00626-4},
  url={https://www.nature.com/articles/s42256-023-00626-4}
}

@inproceedings{hu2022lora,
  title={LoRA: Low-Rank Adaptation of Large Language Models},
  author={Hu, Edward J. and Shen, Yelong and Wallis, Phillip and Allen-Zhu, Zeyuan and Li, Yuanzhi and Wang, Shean and Wang, Lu and Chen, Weizhu},
  booktitle={International Conference on Learning Representations},
  year={2022},
  url={https://openreview.net/forum?id=nZeVKeeFYf9}
}

@inproceedings{houlsby2019adapter,
  title={Parameter-Efficient Transfer Learning for NLP},
  author={Houlsby, Neil and Giurgiu, Andrei and Jastrzebski, Stanislaw and Morrone, Bruna and de Laroussilhe, Quentin and Gesmundo, Andrea and Attariyan, Mona and Gelly, Sylvain},
  booktitle={International Conference on Machine Learning},
  pages={2790--2799},
  year={2019},
  url={https://arxiv.org/abs/1902.00751}
}

@inproceedings{li2021prefix,
  title={Prefix-Tuning: Optimizing Continuous Prompts for Generation},
  author={Li, Xiang Lisa and Liang, Percy},
  booktitle={Proceedings of the 59th Annual Meeting of the Association for Computational Linguistics},
  pages={4582--4597},
  year={2021},
  doi={10.18653/v1/2021.acl-long.353},
  url={https://aclanthology.org/2021.acl-long.353/}
}

@inproceedings{devlin2019bert,
  title     = {{BERT}: Pre-training of Deep Bidirectional Transformers for Language Understanding},
  author    = {Devlin, Jacob and Chang, Ming-Wei and Lee, Kenton and Toutanova, Kristina},
  booktitle = {Proceedings of the 2019 Conference of the North American Chapter of the Association for Computational Linguistics: Human Language Technologies},
  pages     = {4171--4186},
  year      = {2019},
  publisher = {Association for Computational Linguistics},
  doi       = {10.18653/v1/N19-1423},
  url       = {https://aclanthology.org/N19-1423/}
}

@misc{kim2025finetuning,
  title={Fine-Tuning Vision-Language-Action Models: Optimizing Speed and Success},
  author={Kim, Moo Jin and Finn, Chelsea and Liang, Percy},
  year={2025},
  eprint={2502.19645},
  archivePrefix={arXiv},
  primaryClass={cs.RO}
}

@misc{wang2025vlaadapter,
  title={VLA-Adapter: An Effective Paradigm for Tiny-Scale Vision-Language-Action Model},
  author={Wang, Yihao and Ding, Pengxiang and Li, Lingxiao and Cui, Can and Ge, Zirui and Tong, Xinyang and Song, Wenxuan and Zhao, Han and Zhao, Wei and Hou, Pengxu and Huang, Siteng and Tang, Yifan and Wang, Wenhui and Zhang, Ru and Liu, Jianyi and Wang, Donglin},
  year={2025},
  eprint={2509.09372},
  archivePrefix={arXiv},
  primaryClass={cs.RO}
}

@inproceedings{luo2023phc,
  title     = {Perpetual Humanoid Control for Real-time Simulated Avatars},
  author    = {Luo, Zhengyi and Cao, Jinkun and Winkler, Alexander and Kitani, Kris and Xu, Weipeng},
  booktitle = {Proceedings of the IEEE/CVF International Conference on Computer Vision (ICCV)},
  year      = {2023},
  url       = {https://openaccess.thecvf.com/content/ICCV2023/html/Luo_Perpetual_Humanoid_Control_for_Real-time_Simulated_Avatars_ICCV_2023_paper.html},
  arxiv     = {2305.06456}
}

@inproceedings{o2023openx,
  title     = {Open X-Embodiment: Robotic Learning Datasets and RT-X Models},
  author    = {{Open X-Embodiment Collaboration} and O'Neill, Abby and Rehman, Abdul and Gupta, Abhinav and Maddukuri, Abhiram and Gupta, Abhishek and Padalkar, Abhishek and Lee, Abraham and Pooley, Acorn and Gupta, Agrim and others},
  booktitle = {Proceedings of the IEEE International Conference on Robotics and Automation (ICRA)},
  pages     = {6892--6903},
  year      = {2024},
  doi       = {10.1109/ICRA57147.2024.10611477},
  url       = {https://arxiv.org/abs/2310.08864}
}

@inproceedings{octo2024,
  title     = {Octo: An Open-Source Generalist Robot Policy},
  author    = {{Octo Model Team} and Ghosh, Dibya and Walke, Homer and Pertsch, Karl and Black, Kevin and Mees, Oier and Dasari, Sudeep and Hejna, Joey and Kreiman, Tobias and Xu, Charles and Luo, Jianlan and Tan, You Liang and Chen, Lawrence Yunliang and Sanketi, Pannag and Vuong, Quan and Xiao, Ted and Sadigh, Dorsa and Finn, Chelsea and Levine, Sergey},
  booktitle = {Proceedings of Robotics: Science and Systems (RSS)},
  year      = {2024},
  doi       = {10.15607/RSS.2024.XX.090},
  url       = {https://arxiv.org/abs/2405.12213}
}

@article{zeng2025behavior,
  title={Behavior foundation model for humanoid robots},
  author={Zeng, Weishuai and Lu, Shunlin and Yin, Kangning and Niu, Xiaojie and Dai, Minyue and Wang, Jingbo and Pang, Jiangmiao},
  journal={arXiv preprint arXiv:2509.13780},
  year={2025}
}

@article{pan2025agility,
  title={Agility meets stability: Versatile humanoid control with heterogeneous data},
  author={Pan, Yixuan and Qiao, Ruoyi and Chen, Li and Chitta, Kashyap and Pan, Liang and Mai, Haoguang and Bu, Qingwen and Zhao, Hao and Zheng, Cunyuan and Luo, Ping and others},
  journal={arXiv preprint arXiv:2511.17373},
  year={2025}
}

@article{sun2026mosaic,
  title={Mosaic: Bridging the sim-to-real gap in generalist humanoid motion tracking and teleoperation with rapid residual adaptation},
  author={Sun, Zhenguo and Huang, Bo-Sheng and Peng, Yibo and Li, Xukun and Ma, Jingyu and Sun, Yu and Li, Zhe and Jiang, Haojun and Gao, Biao and Bing, Zhenshan and others},
  journal={arXiv preprint arXiv:2602.08594},
  year={2026}
}

@article{wang2026omnixtreme,
  title={Omnixtreme: Breaking the generality barrier in high-dynamic humanoid control},
  author={Wang, Yunshen and Zhu, Shaohang and Zhi, Peiyuan and Li, Yuhan and Li, Jiaxin and Li, Yong-Lu and Xiao, Yuchen and Wang, Xingxing and Jia, Baoxiong and Huang, Siyuan},
  journal={arXiv preprint arXiv:2602.23843},
  year={2026}
}

@article{araujo2025retargeting,
  title={Retargeting matters: General motion retargeting for humanoid motion tracking},
  author={Araujo, Joao Pedro and Ze, Yanjie and Xu, Pei and Wu, Jiajun and Liu, C Karen},
  journal={arXiv preprint arXiv:2510.02252},
  year={2025}
}

@inproceedings{mahmood2019amass,
  title={AMASS: Archive of motion capture as surface shapes},
  author={Mahmood, Naureen and Ghorbani, Nima and Troje, Nikolaus F and Pons-Moll, Gerard and Black, Michael J},
  booktitle={Proceedings of the IEEE/CVF international conference on computer vision},
  pages={5442--5451},
  year={2019}
}

@inproceedings{kim2018prioritized,
  title={Computationally-Robust and Efficient Prioritized Whole-Body Controller with Contact Constraints},
  author={Kim, Donghyun and Lee, Jaemin and Ahn, Junhyeok and Campbell, Orion and Hwang, Hochul and Sentis, Luis},
  booktitle={2018 IEEE/RSJ International Conference on Intelligent Robots and Systems (IROS)},
  pages={1--8},
  year={2018},
  organization={IEEE},
  doi={10.1109/IROS.2018.8593767}
}

@inproceedings{chignoli2021mit,
  title={The MIT Humanoid Robot: Design, Motion Planning, and Control for Acrobatic Behaviors},
  author={Chignoli, Matthew and Kim, Donghyun and Stanger-Jones, Elijah and Kim, Sangbae},
  booktitle={2020 IEEE-RAS 20th International Conference on Humanoid Robots (Humanoids)},
  pages={1--8},
  year={2021},
  organization={IEEE},
  doi={10.1109/HUMANOIDS47582.2021.9555782}
}

@article{zhu2026clot,
  title={CLOT: Closed-Loop Global Motion Tracking for Whole-Body Humanoid Teleoperation},
  author={Zhu, Tengjie and Cai, Guanyu and Zhaohui, Yang and Ren, Guanzhu and Xie, Haohui and Wang, ZiRui and Wu, Junsong and Wang, Jingbo and Yang, Xiaokang and Mu, Yao and others},
  journal={arXiv preprint arXiv:2602.15060},
  year={2026}
}

@article{chen2026holomotion,
  title={HoloMotion-1 Technical Report},
  author={Chen, Maiyue and Wang, Kaihui and Zhang, Bo and Ma, Xihan and Yang, Zhiyuan and Ren, Yi and Huang, Qijun and Zhu, Zihao and Wang, Yucheng and Su, Zhizhong},
  journal={arXiv preprint arXiv:2605.15336},
  year={2026}
}

@misc{unitree_g1,
  title        = {{Unitree G1 Humanoid Robot}},
  author       = {{Unitree Robotics}},
  howpublished = {\url{https://www.unitree.com/g1}},
  year         = {2024},
  note         = {Accessed: 2026-05-22}
}

@misc{unitree_h1,
  title        = {{Unitree H1 Universal Humanoid Robot}},
  author       = {{Unitree Robotics}},
  howpublished = {\url{https://www.unitree.com/h1}},
  year         = {2023},
  note         = {Accessed: 2026-05-22}
}

@misc{limx_oli,
  title        = {{LimX Oli: Full-Size General-Purpose Humanoid Robot}},
  author       = {{LimX Dynamics}},
  howpublished = {\url{https://www.limxdynamics.com/en/products/oli}},
  year         = {2025},
  note         = {Accessed: 2026-05-22}
}

@misc{limx_luna,
  title        = {{LimX Luna Humanoid Robot}},
  author       = {{LimX Dynamics}},
  howpublished = {\url{https://x.com/LimX_Dynamics}},
  year         = {2026},
  note         = {Official product page not yet publicly available at the time of access; accessed: 2026-05-22}
}

@inproceedings{li2023omomo,
  title        = {Object Motion Guided Human Motion Synthesis},
  author       = {Li, Jiaman and Wu, Jiajun and Liu, C. Karen},
  booktitle    = {ACM Transactions on Graphics (SIGGRAPH Asia)},
  year         = {2023}
}

@article{harvey2020lafan,
  title        = {Robust Motion In-betweening},
  author       = {Harvey, F{\'e}lix G. and Yurick, Mike and Nowrouzezahrai, Derek and Pal, Christopher},
  journal      = {ACM Transactions on Graphics (SIGGRAPH)},
  volume       = {39},
  number       = {4},
  year         = {2020}
}

\end{document}